%% file: main.tex
\documentclass[twoside,11pt]{article}

\input{imports}

\begin{document}

\title{Game Plan:\\  What AI can do for Football, and What Football can do for AI}

\author{\name Karl~Tuyls\thanks{Equal contributors.} \thanks{Corresponding author.} , Shayegan~Omidshafiei\footnotemark[1] , Paul~Muller, Zhe~Wang, Jerome~Connor, Daniel~Hennes \\
\addr DeepMind\\
\AND
\name Ian~Graham, William~Spearman, Tim~Waskett, Dafydd Steele \\
\addr Liverpool Football Club\\
\AND
\name Pauline~Luc, Adria~Recasens, Alexandre~Galashov, Gregory~Thornton, Romuald~Elie, Pablo~Sprechmann, Pol~Moreno, Kris~Cao, Marta~Garnelo, Praneet~Dutta, Michal~Valko, Nicolas~Heess, Alex~Bridgland, Julien P\'erolat, Bart De~Vylder, Ali~Eslami, \mbox{Mark~Rowland}, Andrew~Jaegle, Remi~Munos, Trevor~Back, Razia~Ahamed, Simon~Bouton, \\\mbox{Nathalie~Beauguerlange}, Jackson~Broshear, Thore~Graepel, Demis~Hassabis\\
\addr DeepMind}

\maketitle

\begin{abstract}%
\noindent
The rapid progress in artificial intelligence (AI) and machine learning has opened unprecedented analytics possibilities in various team and individual sports, including baseball, basketball, and tennis. More recently, AI techniques have been applied to football, due to a huge increase in data collection by professional teams, increased computational power, and advances in machine learning, with the goal of better addressing new scientific challenges involved in the analysis of both individual players' and coordinated teams' behaviors. The research challenges associated with predictive and prescriptive football analytics require new developments and progress at the intersection of statistical learning, game theory, and computer vision. In this paper, we provide an overarching perspective highlighting how the combination of these fields, in particular, forms a unique microcosm for AI research, while offering mutual benefits for professional teams, spectators, and broadcasters in the years to come. 
We illustrate that this duality makes football analytics a game changer of tremendous value, in terms of not only changing the game of football itself, but also in terms of what this domain can mean for the field of AI. We review the state-of-the-art and exemplify the types of analysis enabled by combining the aforementioned fields, including illustrative examples of counterfactual analysis using predictive models, and the combination of game-theoretic analysis of penalty kicks with statistical learning of player attributes. We conclude by highlighting envisioned downstream impacts, including possibilities for extensions to other sports (real and virtual).
\end{abstract}

\section{Introduction}

\begin{figure}[t!]
    \centering
    \includegraphics[width=0.9\textwidth]{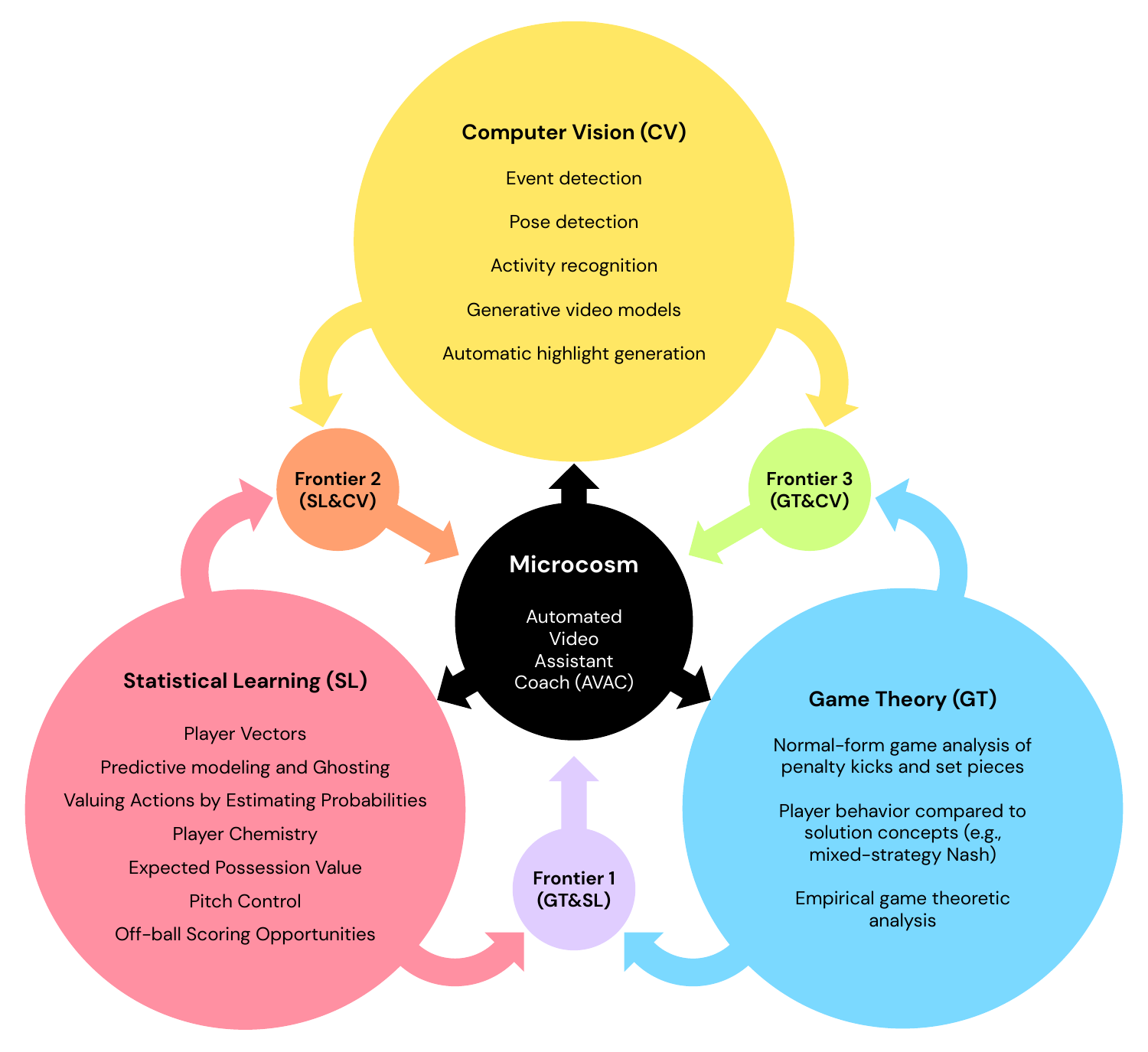}
    \caption{AI research frontiers associated with football analytics. We highlight three foundational areas related to AI research--statistical learning, computer vision, and game theory--which have independently been demonstrated to be effective in analyzing the game of football (with example problems and works from literature shown above per field). We argue that the domain spanned by these research fields is most promising for establishing seminal progress in football analytics in the coming years, along 3 frontiers: \frontierOneGTSL, \frontierTwoSLV, and \frontierThreeGTV. We claim that the above frontiers culminate into a unique microcosm mutually benefiting AI and football analytics.}
    \label{fig:overview}
\end{figure}
Recent years have seen tremendous growing interest in sports analytics, not only from an economic and commercial perspective, but also from a purely scientific one, viz.\,the growing number of publications~\citep{baumer2014sabermetric,shih2017survey,beal_norman_ramchurn_2019} and scientific events organized on the topic (e.g., \citet{mit_sloan_conf}, \citet{cvsports_workshop}, and \citet{kdd_ml_sa}).
As evident in many different downstream domains that have benefited from  applications of artificial intelligence (AI) and machine learning (ML), this is due to important technological advances in data collection and processing capabilities, progress in statistical and in particular {deep learning}, increased compute resources, and ever-growing economic activities associated with sports and culture (e.g., emergent consultancy ventures revolving around sports data collection and statistics~\citep{beal_norman_ramchurn_2019,opta_sports,ChyronHego,InStat,StatsBomb,soccernomics}). 

{Predictive analytics} has been investigated and applied in the context of several sports in the past decades, including basketball~\citep{skinner2010price}, tennis~\citep{walker2001minimax,gauriot2016nash}, and baseball~\citep{Albert02,Lewis04,costa2009practicing,Song17,Puerzer02,Albert10,baumer2014sabermetric}, with data for the latter having been systematically collected since the 19$^{\text{th}}$ century.
Although statistical analysis of data has led to impressive outcomes in various sports (e.g., Moneyball in baseball~\citep{Lewis04,baumer2014sabermetric}), football started participating rather late in this data collection and number-crunching game, with the data science transformation that informs stakeholders (e.g., decisions related to player transfers, scouting, pre- and post-match analysis, etc.) still in its infancy \citep{soccernomics}. 
Several factors influenced this late arrival.
Football takes place under far less controllable settings than other sports due to its outdoor and highly dynamic nature, a larger pitch, a large number of players involved, a low number of player changes and longer non-interrupted game sequences than sports such as basketball.
As a result, football analytics companies have only relatively recently started collecting so-called big data (e.g., high-resolution videos, annotated event-streams, player tracking and pose information). 
Concurrently, only recently have major breakthroughs been made in deep learning, yielding techniques that can handle such new high-dimensional data sets \citep{bengio2009learning,Arel10,lecun2015deeplearning,Schmid15,Goodfellow-et-al-2016}.
Finally, for a long time, credibility in decision-making primarily depended on human specialists such as managers, retired players, and scouts, all of them with track records and experience in professional football, in part due to cultural reasons~\citep{soccernomics,DecroosD19}. 
As a result of these various factors, the potential influence and gains of predictive analytics on the football game have also been less obvious, with sports analytics as a game-changing phenomena not realized until recent years.
In more philosophical terms, \citet{soccernomics} highlight a cultural hesitation regarding the integration of data science into football and an overdependence on gut instincts, noting that ``until very recently, soccer had escaped the Enlightenment". 

Despite football’s late adoption of sports analytics, there are a number of early-bird approaches from different areas of AI such as statistical learning (SL), computer vision (CV), and game theory (GT) that are making initial contributions to support decision-making of managers, coaches and players.
For example, basic statistical learning tools such as principal component analysis (PCA) already enable automated means of identifying player types \citep{DecroosD19}, training of models predicting trajectories of individual teams or imitating league-average behaviors \citep{Le17}, and valuing individual player decisions (such as passes or tackles) in a series of actions leading up to a goal \citep{DecroosBHD19}. 
The study of interactive decision-making as formalized by game theory plays a critical role in AI for systems involving more than one actor (human or artificial).
Game-theoretic tools shed light on players' strategic interactions during scenarios such as penalty kicks, analysis of their decisions in comparison to mathematically-principled baselines, and prediction of their goal-scoring probabilities when playing according to a mixed-strategy Nash equilibrium~\citep{Palacios2003,chiappori2002testing,palacios2016beautiful}.
Enriched with {empirical game theory}~\citep{wellman2006methods, TuylsPLHELSG20,omidshafiei2019alpha} the effects of various high-level strategies pitted against each other can also be analyzed.
Finally, recent developments in computer vision have been employed for player tracking~\citep{Lu13,Liu2013,Bialkowski2015,Gade18}, pose estimation~\citep{Zhang_2019_CVPR,Fastovets,Bridgeman_2019,Sypetkowski19,Sypetkowski}, and automated injury prediction~\citep{Kampakis} based on, e.g., gait and fatigue analysis \citep{Meert18,Ramos20,Claudino,Kakavas,Bartlett}. 

While these separate areas within AI research have independently been demonstrated to be effective for football analytics, we believe that the most pertinent research problems lie in the underexplored intersection of statistical learning, computer vision, and game theory (see \cref{fig:overview}).
Specifically, we pinpoint several frontiers at the intersections of these fields, and identify the ultimate frontier to be the microcosm requiring integrated approaches across all three fields.
A large portion of the state-of-the-art research in football analytics, by contrast, typically falls under the umbrella of one of these areas, with some initial activities taking places at \frontierTwoSLV~\citep{Lu13,Mora17,Mehrasa,Choi_2019_ICCV,Quiroga_2020}, and no notable research activities identified at \frontierOneGTSL and \frontierThreeGTV for football or sports in general. 
At \frontierOneGTSL, game-theoretic analysis is blended with learned predictive models, combining interactive decision-making with predictive modeling to provide more granular analysis tools.
We present a detailed case study illustrating this frontier, revisiting the seminal work of \citet{Palacios2003} on penalty kicks under this new perspective and illustrate how mixing with SL provides deeper insights into penalty-kick taking.
\frontierTwoSLV focuses on research that integrates statistical learning with computer vision, directly learning from video as the primary input and building predictive models (e.g., forecasting player and team behaviors directly).
At \frontierThreeGTV, we classify football research integrating computer vision and game theory, a largely uncharted territory focusing on generative models based on visual inputs, which takes strategic interactions into account.

We claim that the above frontiers culminate into a unique microcosm mutually benefiting both AI and football analytics, to the point it becomes possible to develop, for example, an {Automated Video Assistant Coach} (AVAC).  The AVAC system is an example of what we believe to be the future of human-centric AI research for football, with the aim of integrating all aspects of the frontiers into a cohesive system enabling both understanding and improvement of human football play.
Such an AVAC system is envisioned to improve the overall experience of the game for players, coaches, and spectators alike.

In the following, we first provide an overview of the literature associated with football analytics, subsequently highlighting gaps and sketching a long-term vision for research in the football microcosm.
We lay out the long-term perspective of how AI can benefit the domain of football analytics, and vice versa, by combining the three identified axes in the literature in manners that have not yet been fully explored.
Next, we present illustrative examples that examine penalty-kick taking from a game-theoretic perspective, building on the work of \citet{Palacios2003,palacios2016beautiful}, and bringing several new insights based on data from the main European leagues.
We subsequently demonstrate how this game-theoretic work can be enriched via integration with statistical learning at \frontierOneGTSL, providing several new insights about penalty-kick taking followed by a high-level example of counterfactual trajectory predictions (i.e., a what-if analysis), thus further justifying the football microcosm as a useful AI research domain.

\section{Literature Overview: AI for Football Analytics}
The following sections outline a long-term research vision for AI applied to football analytics. 
We first consider the state-of-the-art with respect to each of the respective areas (GT, SL, and CV) applied to the football domain, after which we look into the opportunities that each of the frontiers brings forth to unlocking larger scientific challenges in football analytics.
The following sections summarize works that lie in these peripheral fields of \cref{fig:overview} and highlight opportunities for further work and added value in each. 

\subsection{Statistical Learning}

Football is arguably the most challenging to analyze of all the major team sports. It involves a large number of players with varied roles, few salient events, and minimal scoring. 
Statistical football analysis attempts to provide quantitative answers to questions that pertain to different aspects of the game. 
Notably, these include the problem of characterizing players' and teams' styles, evaluation of the impact that such teams have on the pitch, and the temporal and counterfactual predictions of players' actions.
When one compares styles of players (and teams, by extension), one usually refers to high level and abstract notions that summarize their unique characteristics.
The goal of the statistical learning line of research is to learn features capturing such information, normally to be used by other down-stream tasks. 
For instance, one means of summarizing information about a football player is through aggregation of their play statistics (e.g., offensive, defensive, or dribbling abilities, shots on target, etc.), and that of their teams~\citep{fernandez2016attacking,stats_playing_styles_2020}.
While such statistics are typically either handcrafted or rely on simple measures of play outcomes, recent works have analyzed them from a statistical learning perspective, using notions such as Player Vectors~\citep{DecroosD19} (detailed in \cref{sec:player_vectors} and analogous techniques in sports such as basketball~\citep{franks2015characterizing}. 
Given the growing success of unsupervised learning methods, there is potential for more advanced representations of player traits to be learned directly from the data. 

In football, it is particularly difficult to assess which players, or groups of players, deserve credit for favorable outcomes. 
For high-scoring team sports (e.g., basketball) one can provide a reasonable answer to this question by restricting attention to actions that have immediate impact on scoring.
By contrast, few goals are scored in football (e.g., 2.72 goals were scored on average per game of the 2019/2020 Premier League season~\citep{pl_goals_2019}).
Consequently, models considering only actions with immediate impact on goals (e.g.\,shots or saves) capture a crucial yet narrow view of the game. 
Moreover, game states in football are significantly more complex than that estimated by current models. 
Features describing them are mostly hand-crafted and only consider on-ball actions. A given pass might be extremely valuable or a poor choice depending on the disposition of the players. 
While these methods are able to value on-ball actions, they rely on sparse signals provided by goals.
Concurrently, off-ball actions significantly impact each game, as exemplified by player actions covering certain pitch areas to prevent attacks, running to open areas to create space for teammates, and so on.
Due to the temporally extended nature of football, the task of inferring the value of actions is an instance of the temporal credit assignment problem in reinforcement learning (RL) \citep{minsky1961steps}. 
The combination of RL techniques with deep learning has great potential to tackle the idiosyncrasies of football and to close the gap between statistical and human analysis. 
It is exciting to see recent progress in this direction in sports analytics, including ice hockey \citep{guiliang2018deep} and football \citep{sun2020cracking, liu2020Deep} (with additional related works detailed in \cref{sec:additional_sl_works}).

Overall, while the above pieces of work showcase the promise of modern statistical methods for temporal predictions in football, this remains an open and challenging problem that will likely require development of novel methods and means of leveraging the diversity of newly-available football data (as later made more precise in terms of what football can do for AI).

\subsection{Game Theory}
Game theory plays an important role in the study of sports, enabling theoretical grounding of players' behavioral strategies.
Numerous works have applied game-theoretic analysis to sports over recent decades~\citep{Sindik2008,Lennartsson15}, including football~\citep{Palacios2003,MOSCHINI2004,Azar2011,Levitt,Buzzachi14,coloma2012penalty,chiappori2002testing}, tennis~\citep{walker2001minimax,gauriot2016nash}, volleyball~\citep{lin2014applying}, basketball~\citep{skinner2010price}, and American football~\citep{emara2014minimax}.
High-level game theory can be applied to team selection and choice of formation.
Set pieces, such as corner kicks and penalties, are particularly amenable to game-theoretic analysis, wherein identified high-level strategies can be pitted against one another and ranked in terms of empirical performance (see \cref{sec:results} for details).
Due to the real-world nature of football, the types of  game-theoretic analysis conducted in this domain have typically been driven by the availability of data sources (in contrast to, e.g., simulation-based domains, where data of varying types and granularity can be synthetically generated).
In particular, the volume of available high-level statistics (e.g., match outcomes spanning across different leagues, seasons, and competition levels) makes football a particularly attractive topic from a behavioral game-theoretic perspective~\citep{camerer2011behavioral,wellman2006methods}. 

From a theoretical perspective, the majority of the existing works exploit the fact that various football scenarios can be modeled as two-player zero-sum games.
For example, in football, the penalty kick situation may be straightforwardly modeled as a two-player asymmetric game, where the kicker's strategies may be neatly categorized as left, center, or right shots.
The controlled nature of these scenarios compounds their appeal from a quantitative analysis perspective;
in the penalty example, the penalty taker, goalkeeper, and ball's initial positions are generally static across all dataset trials (albeit, with minor variations, e.g., scenarios where goalkeepers stand slightly off-center to entice the kicker to shoot in one direction).
In fact, the majority of the literature that analyzes football under a game-theoretic lens focuses on penalty kicks~\citep{Palacios2003,Azar2011,Levitt,Buzzachi14,coloma2012penalty,chiappori2002testing}, which we contend is due to this amenability for analysis via classical game-theoretic solution concepts (such as Nash equilibria).

However until the paradigm shifts away from the classical analysis of set pieces and toward live-play settings (such as those considered by \citet{MOSCHINI2004}), the potential benefits of game-theoretic analysis for football are likely to remain largely untapped.
Moving beyond this classical paradigm involves resolution of significant challenges:
the number of active players in football is quite large (22 including goalkeepers) and the exponential size of the strategy space (with respect to the number of players) makes it more challenging to analyze than games such as tennis or basketball;
the mapping of low-level player actions to strategies is no longer straightforward in durative live plays, due to the variability of player trajectories; 
the duration of plays is also generally longer than sports played in more controlled environments (e.g., tennis), implying that such scenarios may benefit from analysis in the so-called extensive-form, which explicitly consider each player's knowledge, opportunities, and actions, unlike in the simpler (but more feasible to analyze) simultaneous-move, normal-form approaches typically used for set piece analysis (later detailed in \cref{sec:gt_background}).

Nonetheless, the availability of new data types increases the viability of conducting more advanced game-theoretic analysis by bootstrapping to advances in statistical learning and computer vision, as later detailed. 
Surmounting these challenges will benefit football from a game-theoretic perspective, enabling both a descriptive analysis (i.e., understanding the interactions undertaken by players and teams in the presence of others), and a prescriptive one (i.e., suggesting the actions such individuals should have executed).

\subsection{Computer Vision}
Computer vision has seen major breakthroughs in the past decade, thanks to the application of deep learning approaches. 
Progress in tasks such as image classification~\citep{deng2009imagenet}, video action recognition~\citep{Carreira_2017_CVPR}, and pose estimation~\citep{alp2018densepose} have unlocked the possibility of automatically extracting complex information from videos. 
Similarly, in football analytics, computer vision methods can enable enrichment of statistical and game-theoretic approaches, which typically rely on hand-labeled, low-dimensional data.
Video is a particularly appealing signal for football analytics: 
it is a rich data source (likely to contain large amounts of information of interest for decision-making) and cheap to acquire (using only widespread camera sensors).
While raw video frames are insufficiently structured to be processed by traditional rule-based processing techniques, computer vision methods enable the extraction of high level, potentially spatially-detailed representations, ready for use by downstream applications.
Examples of such extraction processes include human pose estimation~\citep{alp2018densepose}, object detection~\citep{ren2015faster}, segmentation~\citep{long2015fully}, tracking~\citep{dong2018triplet}, depth estimation~\citep{godard2017unsupervised}, and event or action detection~\citep{Carreira_2017_CVPR}. 

In addition to explicit, human-interpretable representations, learned counterparts can be used by downstream deep learning models. 
In football, several companies already commercialize tracking information, relating to the players and the ball, automatically extracted from videos recorded by dedicated static cameras, placed to have a view covering the entire terrain~\citep{opta_sports,StatsBomb}. 
Moreover, immersive sports analytics has been gaining popularity with the increased access to specialized hardware and portability~\citep{Lin2020SportsXRI}. 
Computer vision techniques enable reconstruction of real game scenarios, which provide more feedback to players and coaches than 2D screens, and have been extended to other sports media as well~\citep{StriVR}.
Vision-based player tracking, pose estimation, and event detection can improve learning of player skill vectors and predictive modeling, subsequently improving player ranking and game-theoretic analysis of strategies.

While video has been traditionally used as the primary high-dimensional signal for the above applications, other modalities such as audio and text streams can provide extremely rich and complementary information in these settings. 
Audio commentary and news threads readily provide a human interpretation of the events occurring in a scene, which is highly complementary to the spatially fine-grained information available in videos. 
Learning-based approaches for generation of text commentary have been previously investigated \citep{chen2010training}, with temporally-aligned sound and text streams seeing recent application in learning rich representations of video signals~\citep{alayrac2020self,arandjelovic2017look,Miech_2020_CVPR}, having been shown to be useful in a variety of computer vision tasks. 
Although such information significantly overlaps with the structured annotations already used by downstream applications (event annotations, etc.), its unstructured nature enables the presence of a greater information content (e.g., a commentator's tone can provide some cues as to an action's value).

Unique challenges arise when seeking to improve the performance and to enlarge the applications of computer vision methods for broadcast video of football games.
In general, the camera's field of view is centered on the most interesting action in the field, leaving many players out of the frame. 
This poses an interesting geometric problem, as broadcast footage shots typically do not overlap, and many state-of-the-art systems are unable to geometrically relate these multiple shots.
Other challenges arise due to the multiple occlusions of players by one another, further complicating the detection, tracking, and identification tasks. 
These problems can be addressed by geometrical approaches in settings where one can intervene to correct camera positions or, ideally, access cameras' extrinsic and intrinsic parameters (that is, the camera's position, orientation, as well as geometrical and optical properties like its focal length).
However, approaches that do not assume access to such information have the potential to leverage more data.
In contrast with other applications, football broadcast videos present an interesting blend between real, large-scale, complex data and a constrained setting (due to the rules of the game \citep{giancola2018soccernet}), hence providing an attractive setting for developing such approaches.  

\section{Game Plan: Long-term Research Vision}\label{sec:game_plan}
In this section we outline a long-term research vision for football analytics in the microcosm frontiers at the intersection of statistical learning, game theory, and computer vision (see \cref{fig:overview}).
\subsection{Frontier 1: Interactive Decision-making (GT \& SL)}
Learning to make suitable decisions in the presence of other agents is where game theory and statistical learning can converge.
This interdisciplinary area of research has received significant attention in the multi-agent RL community over the past decades, with thorough survey articles~\citep{PanaitL05,ShohamPG07,BusoniuBS08,TuylsW12,BloembergenTHK15,Hernandez-LealK20} available. 
When considering the football domain in particular, it is evident that the potentials of game theory have yet to be fully exploited, let alone its combination with machine learning techniques such as RL.

There are several promising routes in this area that not only are challenging from a football analytics perspective, but also from an AI research one.
Two of these routes are studied further in this article in \cref{sec:results}; one concerns the study of set pieces using the combination of statistical learning with game theory, and a second focuses on predictive modelling for counterfactual analysis.
In the former, which has been mostly studied in a game-theoretic setting, we show how augmenting the analysis with player-specific statistics can provide deeper insights in how various types of players behave or take decisions about their actions in a penalty kick scenario. 
In the latter case, we illustrate how machine learning techniques can facilitate counterfactual analysis in football matches. 
The possibility to predict, for example, trajectories of players can enable investigation of counterfactual scenarios, (e.g., wherein one would like to know how a specific player or team would respond in a specific match scenario).
Doing this enables one to not only learn to generate behaviors, but also leverage game-theoretic techniques for counterfactual analysis.
We defer a more detailed discussion of these research lines to \cref{sec:results}. 

Building on the counterfactual prediction of players' behaviors, one can also consider the possibility of using this as a coaching tool.
Specifically, one can use counterfactual insights to advise tactics to individual players, and even go further by optimizing the overall team strategy depending on the specific opponent in an upcoming match.
This would go beyond the state-of-the-art, which focuses on simply predicting player behaviors;
here, one would seek to actively optimize suggested tactics based on the particular behaviors and play style of the opposing team, and upcoming match-ups.
Such tactical considerations can also be conducted in an iterative manner (e.g., predictions can be made for the opposing team as conditioned on the best-response behavior of the main team), and effective counter strategies can be learned, for instance, using multi-agent RL. 
Such an approach opens the door to a slew of interesting research challenges. For instance, use of multi-agent RL entails definition of a reward function for the players;
while rewarding goals is an obvious candidate, denser reward signals (e.g., associated with successful passes, intercepts, etc.) may be useful for accelerating learning of such policies. 
Such reward functions are also likely to depend on the role of a player in the team and their respective skill set (i.e., may even be heterogeneous across players in a given team), and could also be learned using techniques such as inverse RL~\citep{NgR00}.
Moreover, one may seek to first define an effective `action space' for players (i.e., more granular or structured actions than `move left' or `move right'), before solving the RL problem.
Finally, the combination of the previous learnt models with pitch control \citep{Spearman16}, a technique to determine which player/team has control over a specific area of the pitch, will provide additional information on open space, passing and scoring opportunities, yielding a powerful tool to enable in-match tailored coaching tools.

\subsection{Frontier 2: Predictive Modeling from Videos (SL \& CV)}

Several challenges naturally lie in the frontier between statistical learning and computer vision.
Statistical learning depends on large quantities of labelled data.  Many of the quantities suitable for models of football are the product of hand-labelling data;
on the other hand, vision-based models could automatically identify events which could be fed into such models.
In addition to the quantity of events that vision-based systems could provide, the quality could also be improved (e.g., with events being accurately registered to the corresponding frame, with minimal temporal error compared to human-labeled annotations).

Furthermore, video is a much richer signal compared to what is traditionally used in predictive tasks, such as forecasting future movement of players or predicting the value of individual actions to the game outcome.
The combination of advanced deep learning models and a rich video signal enables learning over subtle clues otherwise not captured in event-stream or tracking data. 
Capturing more of the partially-observable state of a game will ultimately enable more accurate predictions. 
Richer information may additionally help to shed light on the intention of players and thus better address the credit assignment problem in action-value estimation. 

On the other hand, models that better capture the game dynamics may be necessary to resolve some of the limitations of vision-based tracking approaches, which arise as players are occluded or move off camera. 
The resulting ambiguities can likely be resolved using predictive model of player dynamics, which may be trained independently and used as a source of additional inputs to the tracking system or trained as one component of a larger pipeline.
Explicit or implicit access to the game dynamics will very likely also improve vision-based action labeling.
Finally, presenting prediction outcomes by means of synthetic video generation remains an ambitious challenge that combines the task of trajectory prediction with video generation. 
Presenting predictions of counterfactual futures as video will enable intuitive understanding both by coaching personnel and players (e.g., in the same vein as recent work on video generation for tennis matches)~\citep{zhang2020vid2player}.

\subsection{Frontier 3: Generative Game-Theoretic Video Analysis Models (GT \& CV)}

\begin{figure}[t]
    \centering
    \begin{subfigure}[b]{0.32\textwidth}
        \centering
    	\includegraphics[width=\textwidth]{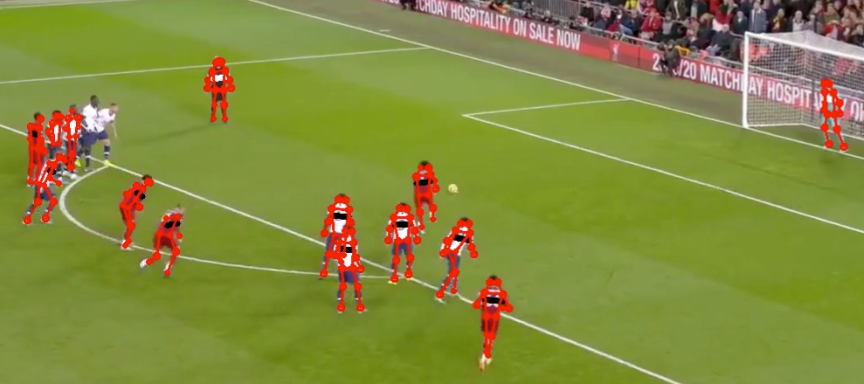}
    \end{subfigure}%
    \hfill%
    \begin{subfigure}[b]{0.32\textwidth}
        \centering
    	\includegraphics[width=\textwidth]{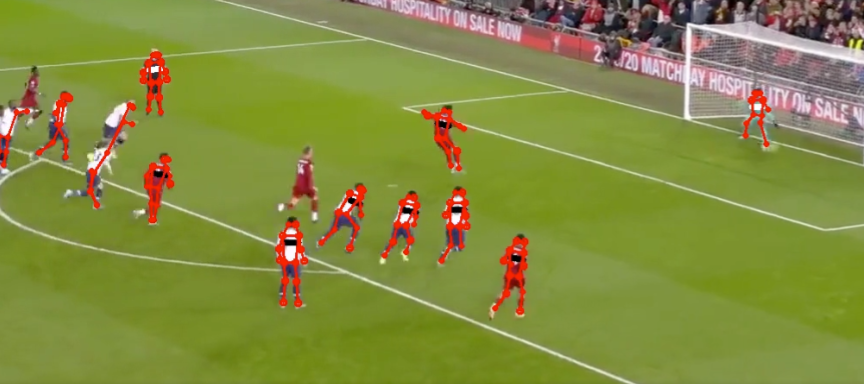}
    \end{subfigure}%
    \hfill%
    \begin{subfigure}[b]{0.32\textwidth}
        \centering
    	\includegraphics[width=\textwidth]{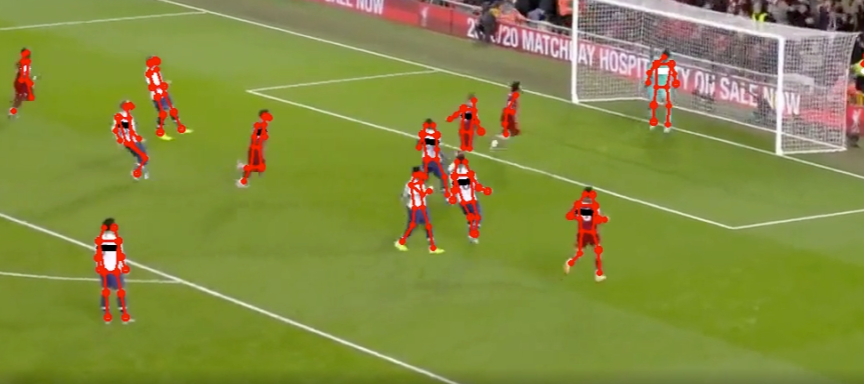}
    \end{subfigure}%
    \caption{Example of a pose estimation model applied to a penalty kick. 
    The results showcased here follow a multi-phase approach~\citep{Papandreou_2017_CVPR} The first stage applies a Faster-RCNN candidate detector~\citep{NIPS2015_5638} which extracts bounding boxes around the players. This is followed by a pose estimation model applied to these bounding box proposals, which identify the player keypoints, and provide the pose estimate.}  
    \label{fig:poses_penalty_kick}
\end{figure}

\begin{figure}[t]
    \centering
    \includegraphics[width=0.80\textwidth,page=1]{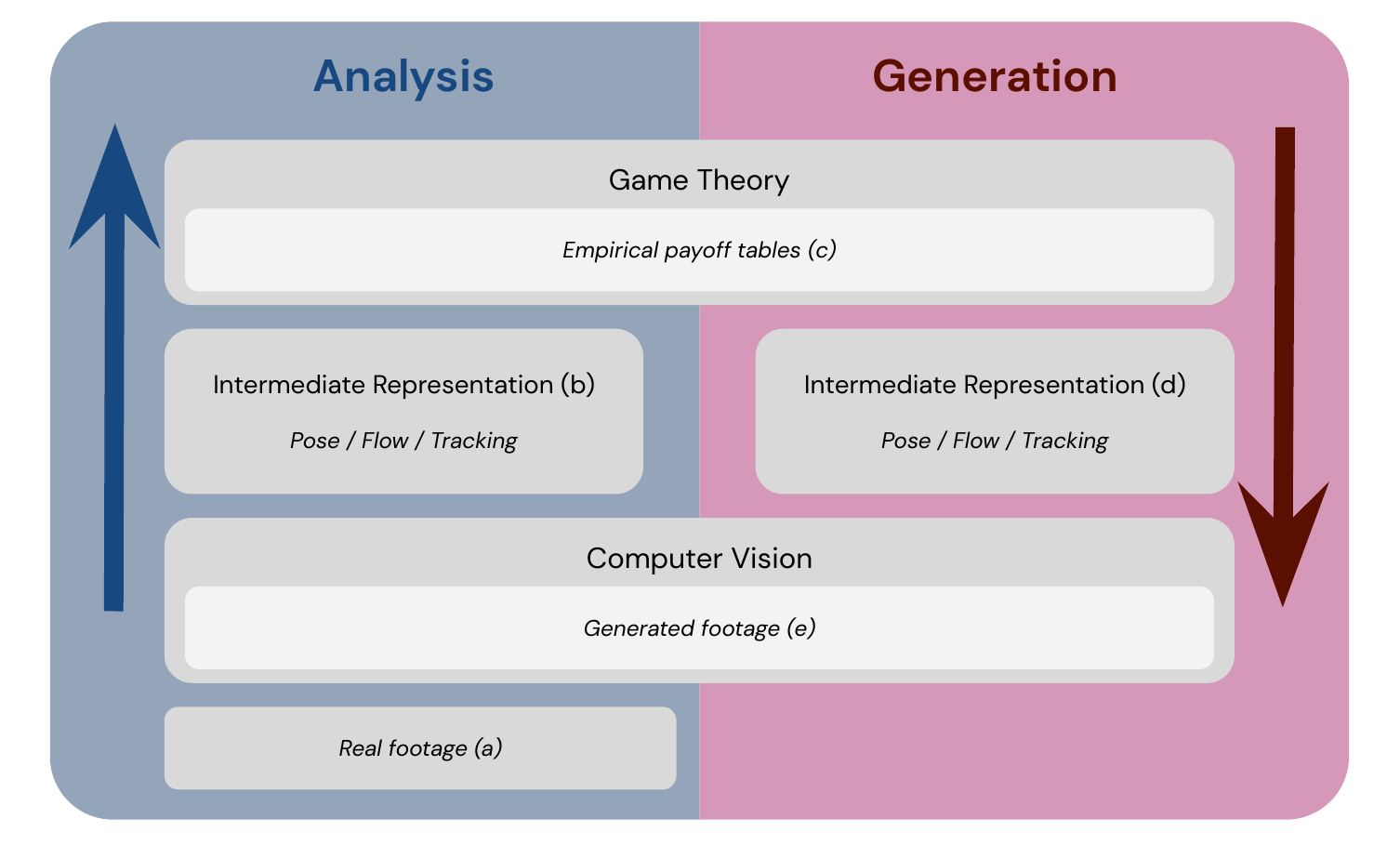}
    \caption{Analysis and generation of video data. From footage video (a), computer vision techniques can extract various representations (b), useful for empirical game-theoretic analysis. In turn, empirical payoff tables (c) can be used in a hierarchical generative process of videos, possibly involving a step in intermediate representation spaces (d). Generated videos (e) could be used for prescriptive analysis, or in the process of capturing the footage videos themselves, closing the cycle between analysis and generation.}
    \label{fig:video_generation}
\end{figure}

Video modeling and game theory can mutually benefit one another. 
In the simplest application, computer vision can provide new features to drive game-theoretic models.
In more advanced applications, game theory can, in turn, guide video generation, as illustrated in \cref{fig:video_generation}.
Throughout a football match, individual players carry out a series of encounters with one another, which can profitably be viewed through game theory. 
Each player may have some idea of the likely success of strategies based on past performance and current in-game observations.
Penalty kicks are an almost idealized example, with a kicker and goalkeeper observing each other for signs of intent, while simultaneously weighing preconceived strategies.
As described previously, computer vision models can be used to automatically extract high-level and potentially spatially-detailed representations that can be complementary to the low-dimensional, hand-collected inputs game-theoretic models typically rely on.
We illustrate this representation extraction pipeline in the left portion of \cref{fig:video_generation}.
In the example of penalty kicks, vision models could identify visual signatures of intent that are difficult to even perceive for humans, let alone annotate;
such information includes, for example, pose estimates extracted from broadcast footage (as visualized in \cref{fig:poses_penalty_kick}, with technical details provided in \cref{sec:pose_estimation}), which can enable inference of the intentions of players, providing valuable insights to improve their respective strategies.

In the reverse direction (right portion of \cref{fig:video_generation}), game-theoretic outputs can be used to improve the plausibility and usefulness of generated videos.
Specifically, generative video models need to precisely capture the data distribution to be useful. 
The game-theoretic context found within football offers an opportunity to constraint such models.
The dynamics of play are complex, with games taking place at the level of opposing individuals to entire teams, though with specific constraints that can inform a generative process.
For example, a hierarchical generative process could condition on high-level latent variables drawn from the distribution described by empirical payoff tables, to represent player decisions conditioning a particular sample.
One could consider an additional hierarchical level of generation targeting intermediate features, possibly building on approaches for predicting future pose~\citep{villegas2017learning, chan2019everybody}, trajectory~\citep{kitani2012activity,bhattacharyya2018long, vu2018memory, bhattacharyya2018bayesian}, action~\citep{vondrick2016anticipating, abu2018will} and shape information~\citep{luc2017predicting, jin2017predicting, luc2018predicting, xu2018structure, sun2019predicting}.

Game-theoretic analysis could be used to assess the plausibility of game simulations.
Specifically, these intermediate outputs could be used directly for prescriptive analysis (e.g., informing players of physical tactics to attempt in a play of interest, based on generated poses) or serve as further conditioning for generation in the raw RGB space.
An alternative direction would be the formulation of new game-theory inspired metrics, for example imposing that the payoff tables extracted from generated samples match those obtained from real data. 
This would be closely related to the Fr\'echet Inception Distance~\citep{heusel2017gans} for generative modeling of images, and the derived Fr\'echet Video Distance~\citep{unterthiner2018towards} for video, which have successfully driven progress in both fields \citep{karras2017progressive, miyato2018spectral, brock2018large, zhang2019self, clark2019adversarial, weissenborn2019scaling, luc2020transformation}.
In turn, such metrics could serve as a basis to design novel losses inspired by game theory.
Besides their potential use for prescriptive analysis, generative models of videos could lead to improvements in broadcast data itself, due to automatic anticipation of the most plausible immediate future outcomes. 
Overall, game-theoretic analysis can inform video generation, in turn enabling influence of the process by which future data will be acquired, hence closing the cycle described in \cref{fig:video_generation}.

\subsection{Football as an AI Testbed: Microcosm}

The development of performative AI algorithms relies on various recurring objectives: learning and acting based on real-world data streams, interpreting the actions of other agents and acting strategically in response, being able to understand and generate natural language for efficient communication with humans, and so on.
As discussed in earlier sections, football analytics involves core problems associated with many of the above AI challenges, though in a more well-controlled scope. 

\begin{figure}[t]
    \centering
    \includegraphics[width=0.95\textwidth,page=1]{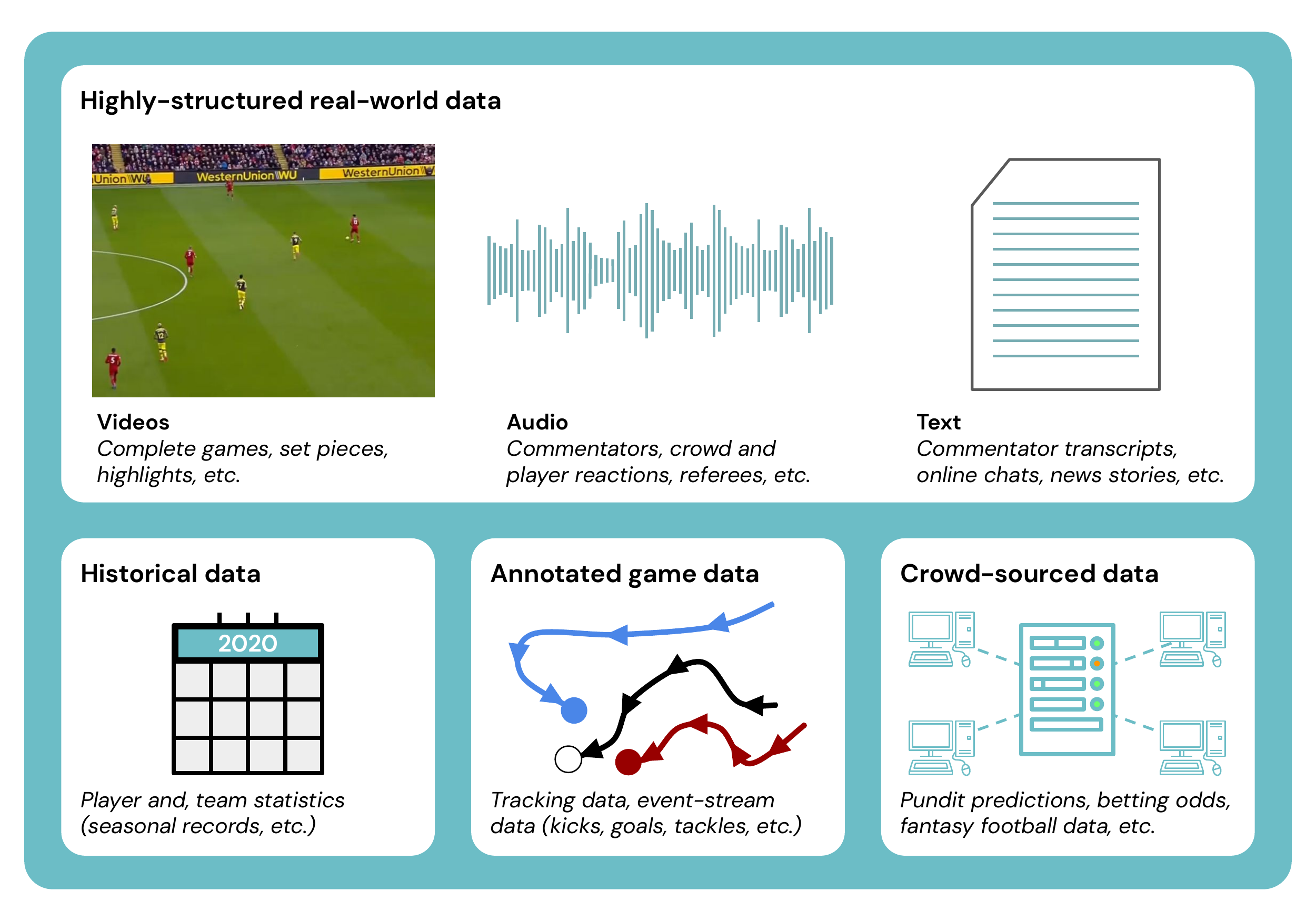}
    \caption{The range of data types relevant to football. At a high-level, highly structured and well-correlated data types are available for each match: video streams, audio streams, and associated text streams. On a more aggregate level, historical play data (e.g., player and team statistics) spanning many decades of play are broadly available for many professional football leagues. Moreover, third-party companies (such as Opta~\citep{opta_sports}, Wyscout~\citep{wyscout}, StatsBomb~\citep{StatsBomb}, and InStat\citep{InStat}) provide annotated player event and tracking data. Finally, there exists a broad range of crowd-sourced data for football, including user preferences and pundit predictions, betting odds, and so on.}
    \label{fig:football_data}
\end{figure}

The value of the football domain for AI can be observed, at a low level, in the range of useful data available for corresponding analytics (see \cref{fig:football_data}).
Real-world data streams such as vision, audio, and text are the mainstay of AI research, and are abundant in football.
Crucially, the various data types available in football are well correlated, in the sense that they typically involve a large amount of shared context--a characteristic researchers can take advantage of. 
For instance:
football video feeds always involve two teams and a ball;
an enormous amount of text is available, yet it is centered on the current game;
the sound of the crowds and commentators can be relied upon to respond to temporally-adjacent events, such as goals and penalties. 
There is a large amount of crowd-sourced data available such as current betting odds and pundits' predictions, a novel form of data atypical in other AI application domains.
Football offers the opportunity for AI to evaluate multi-modal models on synthesized vision, audio, and text data in a unified, though simpler domain than the broader real world. 

Football analytics also currently relies heavily upon hand-labeled data, such as ball and player tracking and identification information. 
This reliance on hand-labeled data imposes a significant barrier for fast-paced analysis, due to the cost and time needed to generate it.
This provides a golden opportunity for AI to accelerate data collection and the subsequent development of novel learning algorithms (by automating the labeling and annotation process) and assisting coaches and decision-makers by instead allowing them to focus their expertise on the tactical analysis of the game itself. 
As such, a worthy long-term challenge for football analytics is to develop such an assistive agent, which uses minimal hand-labeled data: 
an Automated Video Assistant Coach (AVAC). 
A successful AVAC would help players, coaches, and spectators alike. 
Specifically, it could help the players by analyzing their play for weak points to further develop. 
A player's performance throughout a game could be analyzed to suggest improvements in position play and assessing the performance overall. 
Prior to a game, an AVAC could suggest strategies tuned to the opponents of the day. 
Coaches also seek to get the best out of their players, but have a limited amount of time and many players to observe and provide feedback to. 
An AVAC would offer coaches many opportunities to help individual players and the team as a whole, suggesting player rosters for a given game, as well as trading or scouting strategies based on counterfactual evaluation of team performance with brand new players.
Such an AVAC system would have the ability to automatically sift and label huge quantities of video streams, enabling broadcasters and spectators alike to retrieve key moments. 
An AVAC could automatically keep a running tally of information the spectator may find interesting based on their reaction and the current state of play. 
To enhance the spectator experience, the AVAC may automatically generate highlight reels that effectively reflect the flow of the game or summarize the most exciting segments~\citep{zhang2016video,Mahasseni_2017_CVPR,Xiong_2019_CVPR,merler2018excitement,yang2015unsupervised}; 
moreover, the AVAC might suggest related games predicted to engage or interest the spectator. 
For those interested in fantasy football, an AVAC might search for players based on a set of qualities. 
Overall, the possibilities for such an automated system are open-ended. 
To make the research objectives and intermediate benefits of developing an AVAC system concrete, we detail three associated research agendas at increasing levels of abstraction in \cref{sec:footballAI}: representation learning, predictive modeling and decision-making, and human factors. 

\section{Football for AI Research}\label{sec:footballAI}

\begin{figure}[t]
    \centering
    \includegraphics[width=\textwidth,page=1]{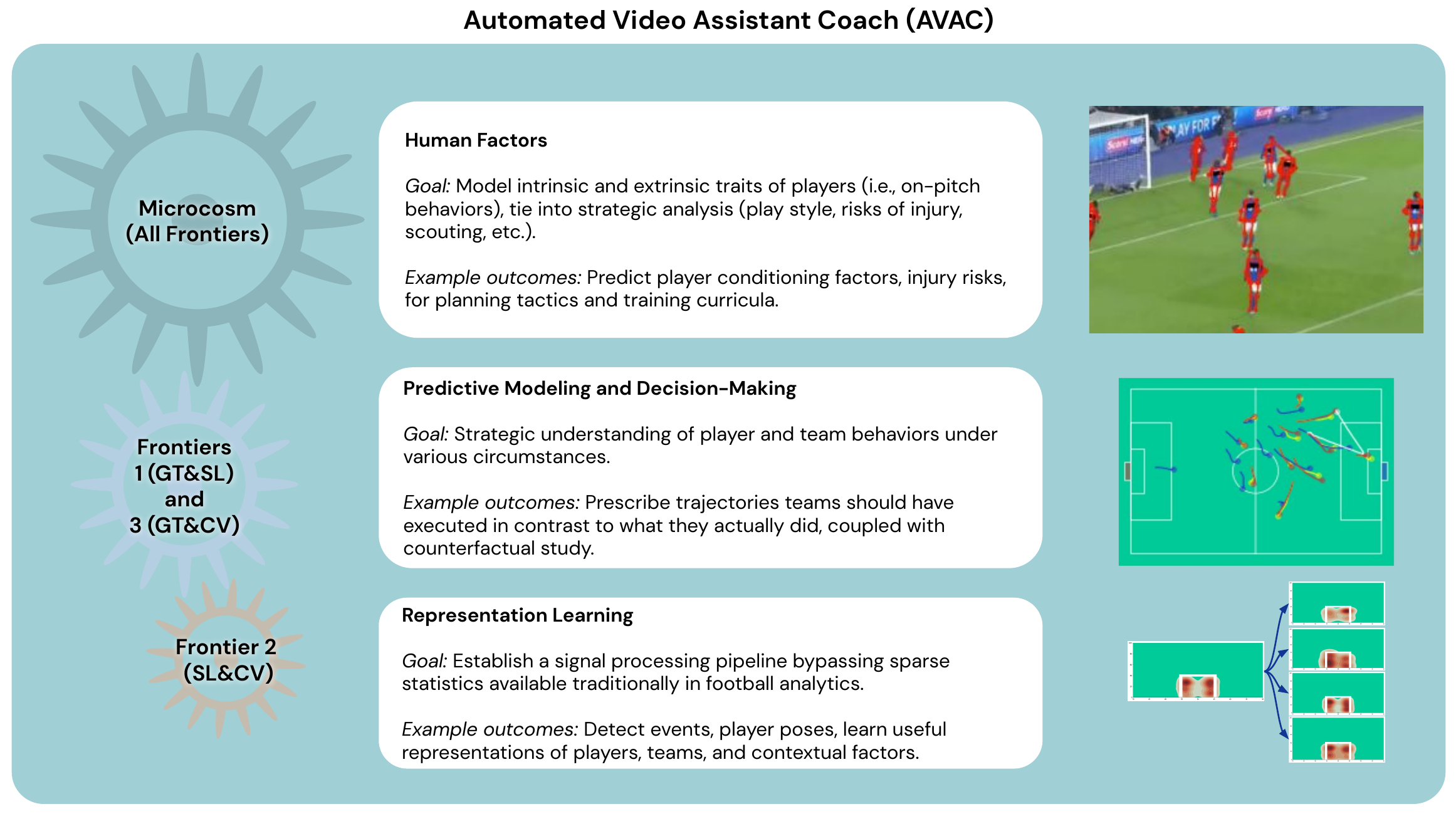}
    \caption{Hierarchy of key research challenges, defined over three layers. 
    The foundational layer targets representation learning over the various input modalities available in football to generate useful representations for more complex learning tasks targeted in the subsequent layers:
    {prescriptive and predictive analysis} and {modeling of human factors}.
    }
    \label{fig:layers}
\end{figure}

We next consider the dual perspective of the potential unique challenges and opportunities associated with the football domain that make it an interesting testbed for AI research. 
We introduce here a hierarchy of key challenges associated with football research, illustrated in \cref{fig:layers}, as defined over three layers:
the foundational layer concerns  representation learning, operating directly on the various input modalities available in football (e.g., time-synchronized videos, event-streams, and tracking data) to generate useful representations for the more complex learning tasks targeted in the subsequent layers of prescriptive and predictive analysis, and modeling of human factors.
We next detail each of these research layers, drawing connections with the three aforementioned frontiers.

\subsection{Representation Learning}
The variety of hand-labeled football statistics make for natural fodder for machine learning algorithms.
These algorithms range from classification and regression tools (e.g., in expected possession value models~\citep{fernandez2019decomposing}), generative models (e.g., in trajectory generation models~\citep{Le17,LeY0L17,yeh2019diverse,li2020generative}), and variational auto-encoding models (player embeddings).
The success of machine learning algorithms generally depends on data representation, as different representations can entangle and obfuscate various explanatory factors of variation behind low-level sensory data.
In football analytics, although expert knowledge is widely used to help design existing representations, learning with generic priors bears promise for avoiding such hand-encoded knowledge. 
Under this view, we identify three unique challenges related to representation learning, detailed next.

The first challenge concerns learning representations with multi-modal football data.
Particularly, in football analytics, it remains a fundamental challenge to effectively recognize {long-duration} playing styles of individual players and teams given the variety of data types available (as detailed earlier).
While expert knowledge goes a long way towards analyzing these multi-modal data sources, it remains insufficient to process them efficiently.
The increasing multitude of input modalities available to football analysts are likely to challenge existing methods of representation learning, thus driving researchers to develop cohesive models that take these many modalities into account simultaneously.

The second challenge concerns learning contextual representations of individual players.
Due to the dynamics and uncertainty of football outcomes, long-term static representations for predictive modeling of in-game events are likely to be beneficial when used in conjunction with representations of individual players.
For example, a player passing the ball may take into account the context of the game to estimate the most appropriate receiver that maximizes the probability of scoring.
Another concrete example is using contextual representations to identify the dynamic roles of players, which may change given the game context and must be inferred and controlled to tactically counter the opposing team.
Finally, player behaviors depend not only on the game context, but also on their and the opposing team's overall strategy (e.g., formations, tactical advice provided by the coaching staff, etc.).
    
Finally, in addition to learning representations of individual players, identifying an effective means of contextualizing or ranking entire teams is another unresolved challenge. 
Teams are usually ranked with historical match results and the collective performance of individual players, which can be coarse (i.e., may not reveal the long-term playing styles of teams) and may fail to reflect in-game dynamics. 
Overall, to tackle these challenges, we aim to achieve two goals: i) learning representations that are able to characterize the long-term playing styles of football teams, ii) learning contextual representations of football teams that are able to depict in-game dynamics.

\subsection{Predictive Modeling and Decision-Making}

Learning useful representations (i.e., as opposed to hand-coded features) serves as an important means of advancing subsequent predictive and prescriptive analysis of football matches.
Specifically, dense embeddings that summarize not only the state of a particular game, but also historical trends evident throughout many games (e.g., across seasons) will serve as enablers of the more accurate, impactful, and longer-horizon predictions of match outcomes.
The interaction between predictive-prescriptive models is envisioned to be tightly-linked with game-theoretic analysis, thus coupling this direction of research most closely with \frontierThreeGTV and \frontierOneGTSL (see \cref{fig:layers}).

The combination of these fields with game theory is likely to usher in new opportunities for coaches and decision-makers.
For example, predictive models of football players at the trajectory-level~\citep{Le17,LeY0L17,sun2019stochastic} currently treat the game as a black-box dynamical process (i.e., a system of dynamic entities making decisions solely based on the joint on-pitch state of the teams);
such models do not yet account for the game-theoretically driven counterfactual responses of players to one another (e.g., taking into account the current game score, time remaining, relative strength of the two teams, impact of current game decisions on upcoming matches, etc.).
Conducting such an analysis of these models involves identification of high-level strategies typically used by empirical game-theoretic techniques (so-called meta-strategies)~\citep{wellman2006methods,TuylsPLHELSG20}.
These meta-strategies, for example, could be clusters of on-pitch behaviors correlated with play style, counterattack types, defense schemes (such as zonal vs. man-to-man defense), and so on.
While such meta-strategies are typically manually defined, automatically learning them poses an interesting challenge.
Appropriate clustering and identification of such meta-strategies involves not only access to a large, representative dataset of plays, but also the aforementioned learned representations that summarize the most relevant context for game theory models.

Synthesis of empirical games over the possible meta-strategies of two opposing teams can be used to forecast the performance of various team tactics when pitted against one another (e.g., investigating the Nash equilibrium of football at a higher level, rather than the typically considered low-level scenarios such as penalty kicks).
Moreover, while characterization and ranking of players has received considerable attention in the literature~\citep{DecroosD19,decroos2019actions,bransen2020chemistry}, automated ranking of {tactics} has received considerably less attention~\citep{decroos2018automatic,meerhoff2019exploring}.
Application of game-theoretic analysis techniques here remains unexplored to the best of our knowledge.
Analysis of empirical games using meta-strategies conditioned on player identities would be beneficial for counterfactually evaluating player performance in new teams (i.e., for scouting). 
For training staff, a model that enables accurate evaluation of players' contributions to the team's overall strategy would be valuable, for example, for pinpointing which players to coach or to substitute.
For broadcasters, automatic identification of salient, exciting meta-strategies (e.g., those that are typically low in probability yet high in payoff, or games where there is a large difference in terms of the play styles or meta-strategies of the teams) can be used for automatic generation of highlight reels. 

Learning the appropriate meta-strategies and associated predictive models are, simultaneously, challenging in football due to the number of players involved on-pitch (and the exponential size of the strategy space with respect to this quantity).
Despite this, the development of richer models leveraging more complex input modalities (e.g., video-based representations) is likely to unlock commensurate benefits (in terms of improved predictions and strategic decision-making) for football analysts. 

\subsection{Human Factors}

The human-centric nature of football analytics stems from several factors: 
coaching and improvement of individual play and coordinated team play through predictive and prescriptive modelling, injury and fatigue prediction, and psychological analysis of players. 
This focus distinguishes it sharply from, for example, challenges such as RoboCup~\citep{RoboCup}.
In contrast to the robot-centric focus of RoboCup (which revolves around developing robotic footballing agents~\citep{Visser,AB05,LNAI09-kalyanakrishnan-1,AAMAS11-urieli,ICLR16-hausknecht,AAAI17-Hanna}), the focus in football analytics is entirely on understanding and improving human gameplay and team coordination based on an integrated approach from the three research areas involved. 
Another key difference concerns evaluation of the impact of said analysis on human play, which is distinct from evaluation of robotic agents in the RoboCup project.
Namely, human play is significantly more difficult to realistically simulate and systematically evaluate (in contrast to evaluation on robotics platforms).
Moreover, the football analytics frontiers targeted here entail taking into account human factors such as injury and fatigue, but also inter-player relationships and their effects on play efficiency and cooperation, psychological challenges such as pressure or mental state, notably on recent transfers, and their impact on play performance, and overall player discipline and tendency to follow the best plan for the team instead of the best plan for themselves.

Injury prediction is another topic of interest. 
Injury prediction is the task of predicting the probability that a player will sustain an injury given data on the past and current seasons. 
Previous studies have investigated the acute-chronic workload ratio (ACWR) as a predictor for sports-related muscular injuries~\citep{gabbett2010development}.
Acute workload measures an athlete's workload over 1 week, while chronic workload is the average workload over the past 4 weeks.
\citet{Rossi} use richer workload measures extracted from Electronic Performance and Tracking Systems (EPTS) to train a decision tree to predict the probability of future injuries. 
Their approach uses manually designed features that aim to temporally aggregate players' workload histories. 
Recent work uses Convolutional Neural Networks (CNNs) applied to EPTS time-series data directly, thereby alleviating the need for hand-designed time-aggregated features~\citep{gabbett2010development}. Current injury prediction methods are limited to a binary signal and do not explicitly capture uncertainty of the prediction, the type of injury, the severity, nor projected time to recover. Progress in this direction is likely limited by the availability of data; preventive measures are taken by sports clubs and as a result injuries occur relatively infrequently, although accurate prediction of such injuries (or determination of whether current performance sensors are sufficient for doing so) is a promising avenue for application of AI techniques.

\section{Illustrative Examples: \frontierOneGTSL}\label{sec:results}

In this section, we highlight some of the benefits that the combination of frontiers can yield for football analytics. 
We focus these examples on \frontierOneGTSL, in particular, given the track record of game theory and statistical learning work done in football analytics in recent years; 
thus, this section provides a concrete sampling of the types of multi-disciplinary contributions that can be made via the proposed microcosm-centric vision.
In the following, we first provide an overview of the necessary game theory background.
Subsequently, we conduct an in-depth analysis of real-world football data under the lens of \frontierOneGTSL, providing new insights into penalty kick scenarios by combining statistical learning with game theory. 

\subsection{Game Theory: Elementary Concepts}\label{sec:gt_background}

Empirical game theory has become an important tool for analysis of large-scale multi-agent settings, wherein either a large amount of data involving agent interactions is readily available, or is collected through simulations of the system under study for the construction of the games \citep{wellman2006methods,TuylsPLHELSG20}.
Empirical game-theoretic modeling of penalty kick taking and set pieces facilitates strategic understanding of player and team behavior under various circumstances (e.g., play according to a Nash equilibrium), and can assist  both in predicting opponent behavior and prescribing how a player (or team) should behave in the presence of other players (teams). 
These game-theoretic models can be leveraged in pre- and post-match analysis, and can be combined with analysis of dynamic trajectory behavior (e.g., generative trajectory prediction or `ghosting', as later described). 
Additionally, the models can be enriched by carrying out Empirical Game Theoretic Analysis (EGTA) on meta-game models of set pieces, automatically clustering and identifying useful meta-strategies, and providing insights into higher-level team strategies.

A common representation of a game used for EGTA analysis is a Normal Form Game (NFG), defined in the following.
\begin{definition}[Normal Form Games (NFG)]
    A game $G = \langle P, (S_i), (u_i) \rangle $ consists of a finite set of players, $P$, indexed by $i$; a nonempty set of strategies $S_i$ for each player; and a utility function $u_i : \times_{j\in P} S_j \rightarrow \mathbb{R}$ for each player.
\end{definition}
In this work, we solely focus on bimatrix games, which are 2-player NFGs, $G = \langle P, (S_1,S_2), (u_1,u_2) \rangle $ with $|P|=2$.
The utility functions $(u_1,u_2)$ can be described in terms of two payoff matrices $A$ and $B$, wherein one player acts as the row player and the other as the column player. 
Both players execute their actions simultaneously.
The payoffs for both players are represented by bimatrix $(A, B)$, which gives the payoff for the row player in $A$, and the column player in $B$ (see \cref{fig:bimatrix} for a two strategy example). 

\begin{figure}[tb]
  \centering
  \gamematrix{}{}{a_{11}, b_{11}}{a_{12},b_{12}}{a_{21}, b_{21}}{a_{22}, b_{22}}
  \caption{Bimatrix payoff tables $(A, B)$ for a two-player, two-action NFG.
    Here, the row player chooses one of the two rows, the column player chooses on of the columns, and the outcome of their joint action determines the payoff to each player.
  }
  \label{fig:bimatrix}
\end{figure}

A player~$i$ may play a {pure strategy}, $s_i\in S_i$, or a {mixed strategy}, $\sigma_i\in\Delta(S_i)$, which is a probability distribution over the pure strategies in $S_i$.
In a {strategy profile} $\sigma=(\sigma_1,\sigma_2)$, each player has a strategy $\sigma_i$.
We use notation $\sigma_{-i}$ to denote a strategy profile for all players excluding~$i$.
Having defined NFGs, we can model empirical games as an NFG wherein player payoffs are directly computed from data of real-world interactions or simulations. 
For example, one can construct a win-loss table between two chess players when they both have access to various strategies.

Given an NFG, the traditional solution concept used in game theory is the Nash equilibrium, which selects strategy profiles such that no player can benefit from unilateral deviation: 
\begin{definition}[Nash Equilibrium]
    A strategy profile $\sigma^*$ is a Nash equilibrium if and only if,
    \begin{equation}
        \E [u_i(\sigma_i^*,\sigma_{-i}^*)] \geq \E [u_i(s'_i,\sigma_{-i}^*)] \qquad \forall i \quad \forall s'_i\in S_i. 
    \end{equation}
\end{definition}

In a so-called $\epsilon$-Nash equilibrium, there exists at least one player who could gain by deviating to another strategy, but that gain is bounded by $\epsilon$. 
More formally:
\begin{definition}[$\epsilon$-Nash Equilibrium]
    A strategy profile $\sigma^*$ is an $\epsilon$-Nash equilibrium if and only if there exists $\epsilon > 0$ such that,
    \begin{equation}
         \E [u_i(\sigma_i^*,\sigma_{-i}^*)] \geq \E [u_i(s'_i,\sigma_{-i}^*)] - \epsilon \qquad \forall i \quad \forall s'_i\in S_i.
    \end{equation}
\end{definition}

\subsection{Game Theory for Penalty Kick Analysis}\label{sec:egta_results}

\begin{figure}[t]
    \centering
    \includegraphics[width=0.5 \textwidth]{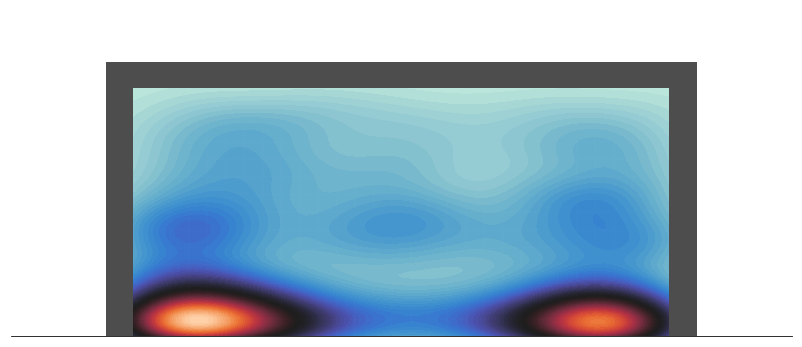}
    \caption{Visualization of shot distribution for the penalty kicks in the considered dataset.}
    \label{fig:shot_distribution}
\end{figure}

\begin{figure}[t]
    \centering
    \includegraphics[width=0.65\textwidth]{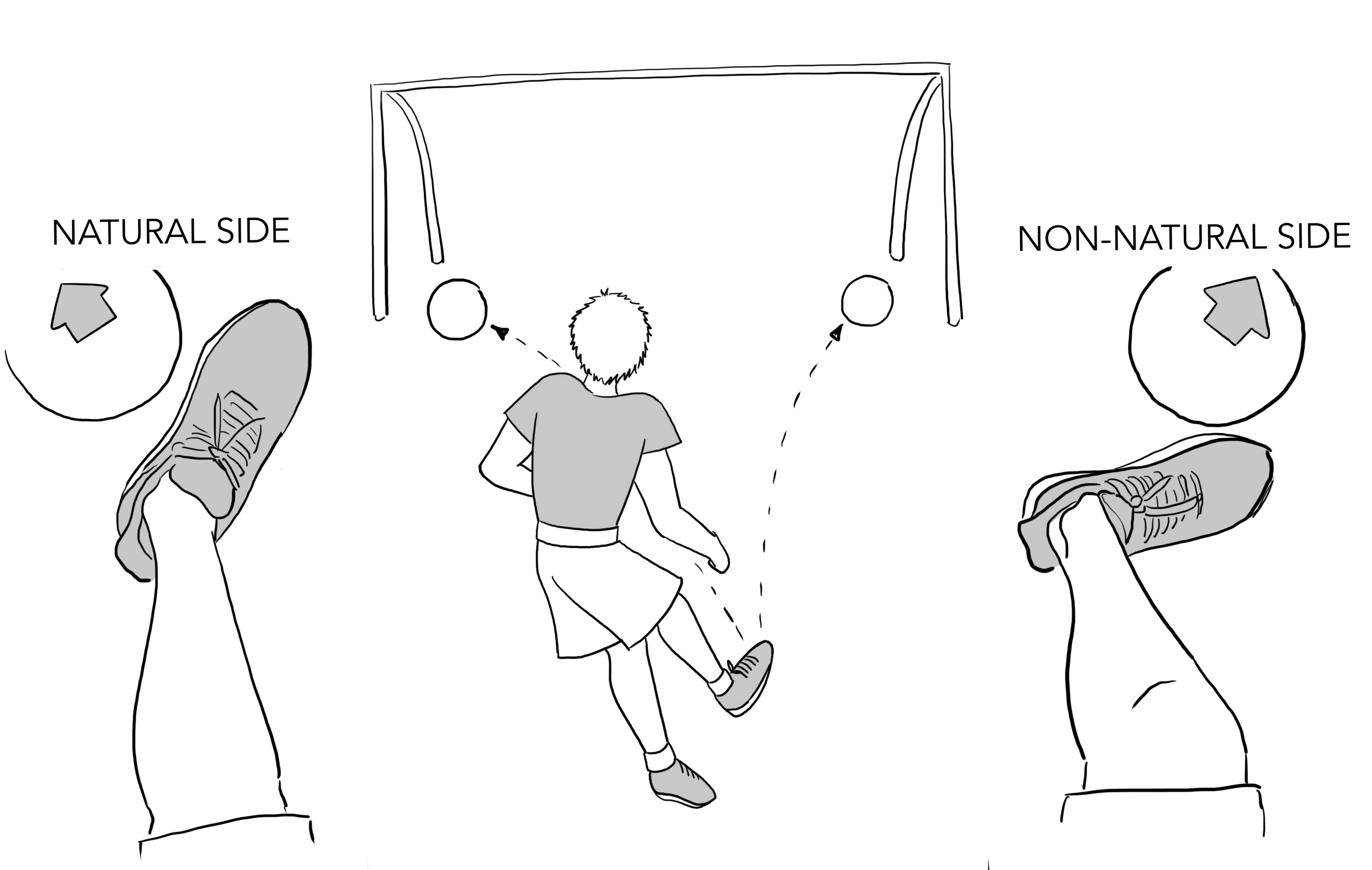}
    \caption{Illustration of natural vs. non-natural sides.}
    \label{fig:natsides}
\end{figure}

\begin{table}[t]
    \centering
    \caption{Penalty kick distribution over leagues considered (12399 kicks in total).}
    \begin{subtable}{.5\textwidth}
    \begin{tabular}{rr}
        \toprule
        League & \# Kicks  \\
        \midrule
        Italian Serie A 	 & 607 	 \\
        US Major League Soccer 	 & 575 	 \\
        English Npower Championship 	 & 569 	 \\
        Spanish Segunda Division 	 & 568 	 \\
        Spanish La Liga 	 & 531 	 \\
        French Ligue 1 	 & 497 	 \\
        German DFB Pokal 	 & 441 	 \\
        Brazilian Série A 	 & 440 	 \\
        English Barclays Premier League 	 & 436 	 \\
        German Bundesliga 	 & 409 	 \\
        Dutch Eredivisie 	 & 398 	 \\
        German Bundesliga Zwei 	 & 389 	 \\
        Portuguese Primiera Liga 	 & 352 	 \\
        Saudi Arabian Profess. League 	 & 337 	 \\
        Russian Premier League 	 & 329 	 \\
        Chinese Super League 	 & 324 	 \\
        Copa Libertadores 	 & 322 	 \\
        Belgian Jupiler League 	 & 287 	 \\
        Turkish Super Lig 	 & 284 	 \\
        French Ligue 2 	 & 270 	 \\
        Argentina Primera (Anual) 	 & 261 	 \\
        English Capital One Cup 	 & 234 	 \\
        Mexican Primera (Clausura) 	 & 234 	 \\
        Colombia Primera Apertura 	 & 221 	 \\
        Norwegian Tippeligaen 	 & 219 	 \\
        AFC Champions League 	 & 193 	 \\
        International Champions Cup 	 & 188 	 \\
        Australian A-League 	 & 172 	 \\
        Copa Chile 	 & 172 	 \\
        English FA Cup 	 & 153 	 \\
        Copa do Brasil 	 & 153 	 \\
    \bottomrule
    \end{tabular}
    \end{subtable}%
    \begin{subtable}{.5\textwidth}
    \begin{tabular}{rr}
    \toprule
    League & \# Kicks  \\
    \midrule
        Chile Primera (Apertura) 	 & 151 	 \\
        Japanese J-League 	 & 149 	 \\
        English League 1 	 & 139 	 \\
        English League 2 	 & 130 	 \\
        Austrian Bundesliga 	 & 129 	 \\
        Danish Superligaen 	 & 115 	 \\
        European World Cup Qualifiers 	 & 108 	 \\
        Internationals 	 & 93 	 \\
        African Cup of Nations 	 & 90 	 \\
        United Soccer League 	 & 80 	 \\
        European Championship Qualifiers 	 & 78 	 \\
        Swedish Allsvenskan 	 & 74 	 \\
        Coppa Italia 	 & 67 	 \\
        Copa America 	 & 51 	 \\
        FIFA Club World Cup 	 & 51 	 \\
        World Cup 	 & 48 	 \\
        European Championship Finals 	 & 45 	 \\
        Champions League Qualifying 	 & 41 	 \\
        Confederations Cup 	 & 39 	 \\
        UEFA Europa League Qualifying 	 & 32 	 \\
        Coupe de France 	 & 29 	 \\
        Belgian UEFA Europa League Play-offs 	 & 24 	 \\
        German 3rd Liga 	 & 23 	 \\
        Russian Relegation Play-offs 	 & 15 	 \\
        Dutch Relegation Play-offs 	 & 13 	 \\
        Copa Sudamericana 	 & 9 	 \\
        Friendly 	 & 4 	 \\
        German Bundesliga Playoff 	 & 3 	 \\
        German Bundesliga 2 Playoff 	 & 3 	 \\
        Swedish Relegation Play-off 	 & 1 	 \\
        & \\ 
        \bottomrule
    \end{tabular}
    \end{subtable}
    \label{tab:leaguesdistribution}
\end{table}

For our analysis we use a data set of $12399$ penalty kicks based on Opta data~\citep{opta_sports}. In \cref{fig:shot_distribution} we show a heatmap of the shot distribution of the penalty kicks in our data set. \Cref{tab:leaguesdistribution} shows the distribution of the penalty kicks over the various leagues we consider.

\citet{Palacios2003} examines penalty kick scenarios from a game-theoretic perspective, using empirical payoff tables to determine whether the associated kickers and goalkeepers play a Nash equilibrium. 
Here we revisit the work of \citet{Palacios2003}, by first reproducing several of its key results with a substantially larger and more recent data set from the main professional football leagues in Europe, Americas, and Asia (for comparison, the data set used in the work of \citet{Palacios2003} consists of 1417 penalty kicks from the 1995-2000 period, whereas ours contains $12399$ kicks from the 2011-2017 period). 
While several results of this earlier work are corroborated, we also find surprising new additional insights under our larger dataset.
We then go further to extend this analysis by considering larger empirical games (involving more action choices for both kick-takers and goalkeepers). 
Finally, we develop a technique for illustrating substantive differences in various kickers' penalty styles, by combining empirical game-theoretic analysis with Player Vectors~\citep{DecroosD19} illustrating the added value and novel insights research at \frontierOneGTSL of the microcosm can bring to football analytics.

\begin{table}[t]
    \centering
    \caption{Natural (N) / Non-Natural (NN) payoff tables for Shots (S) and Goalkeepers (G).}
    \begin{subtable}[b]{0.48\textwidth}
        \centering
    	\caption{\citet{Palacios2003} payoff table.}
    	\label{tab:Palacios}
        \begin{tabular}{rRR}
            \toprule
            {} & N-G & NN-G \EndTableHeader \\
            \midrule
            N-S     & 0.670  & 0.929 \\
            NN-S &   0.950 & 0.583 \\
            \bottomrule
        \end{tabular}
    \end{subtable}
    \hfill
    \begin{subtable}[b]{0.48\textwidth}
        \centering
        \caption{Reproduced table.}
        \label{tab:OurversionofPalacios}
        \begin{tabular}{rRR}
            \toprule
            {} & N-G & NN-G \EndTableHeader \\
            \midrule
            N-S     &  0.704    &   0.907 \\
            NN-S    &  0.894    &   0.640 \\
            \bottomrule
        \end{tabular}
    \end{subtable}\\
    \par\bigskip
    \begin{subtable}[b]{0.48\textwidth}
        \centering
        \caption{\citet{Palacios2003} Nash probabilities.}
        \label{tab:PalaciosNash}
        \begin{tabular}{rMM|MM}
            \toprule
            {} &       NN-S &      N-S &       NN-G & N-G \EndTableHeader \\
            \midrule
            Nash  &  0.393 & 0.607 & 0.432 & 0.568  \\
            Empirical  &  0.423 &  0.577 &  0.400 & 0.600 \\
            \bottomrule
        \end{tabular}
        \setlength{\fboxrule}{0pt}
        \fbox{Jensen–Shannon divergence: 0.049\%}
    \end{subtable}
    \hfill
    \begin{subtable}[b]{0.48\textwidth}
        \centering
        \caption{Reproduced table Nash probabilities.}
        \label{tab:OurPalaciosNash}
        \begin{tabular}{rMM|MM}
            \toprule
            {} &       NN-S &      N-S &       NN-G & N-G \EndTableHeader \\
            \midrule
            Nash       &  0.431 &  0.569 &  0.408 &                0.592 \\
            Empirical  &  0.475 &  0.525 &  0.385 &                0.615 \\
            \bottomrule
        \end{tabular}
        \setlength{\fboxrule}{0pt}
        \fbox{Jensen–Shannon divergence: 0.087\%}
    \end{subtable}
\end{table}

As in \citet{Palacios2003}'s work, we first synthesize a 2-player 2-action empirical game based on our penalty kick data set.
\Cref{tab:Palacios} illustrates the $2\times2$ normal form as presented by \citet{Palacios2003}.
The actions for the two players, the kicker and goalkeeper, are respectively visualized in the rows and columns of the corresponding payoff tables, and are detailed below.
The respective payoffs in each cell of the payoff table indicate the win-rate or probability of success for the kicker (i.e., a score); 
for ease of comparison between various payoff tables, cells are color-graded in proportion to their associated values (the higher the scoring probability, the darker shade of green used).

The choice of player actions considered has an important bearing on the conclusions drawn via empirical game-theoretic analysis. 
The actions used by \citet{Palacios2003} in \cref{tab:Palacios} correspond to taking a shot to the natural (N) or non-natural (NN) side for the kicker, and analogously diving to the natural side or non-natural side for the goalkeeper. 
\Cref{fig:natsides} provides a visual definition of natural versus non-natural sides.
Specifically, as players tend to kick with the inside of their feet, it is easier, for example, for a left-footed player to kick towards the right (from their perspective);
thus, this is referred to as their natural side.
Analogously, the natural side for a right-footed kicker is to kick towards their left. The natural side for a goalkeeper depends on the kicker in front of him. 
Specifically, when facing right-footed kickers, goalkeepers' natural side is designated to be their right; 
vice versa, when they face a left-footed kicker, their natural side is to their left. 
Importantly, shots to the center count as shots to the natural side of the kicker, because, as explained in \citet{Palacios2003}, kicking to the center is considered equally natural as kicking to the natural side by professional football players~\citep{Palacios2003}. 

\Cref{tab:OurversionofPalacios} shows our reproduction of \cref{tab:Palacios} of \citet{Palacios2003}, computed using $12399$ penalty kicks spanning the aforementioned leagues in our Opta-based dataset; importantly, players (goalkeepers and kickers) appear at least 20 times each in this dataset, to ensure consistency with \citet{Palacios2003}. 
The trends in these two tables are in agreement: 
when the goalkeeper and the kicker do not choose the same sides of the goal, shot success rate is high; 
otherwise, when the keeper goes to the same side as the kicker, success rate is higher for natural shots than for non-natural shots. 
We also include Nash and empirical probabilities for \citeauthor{Palacios2003}'s dataset and ours, respectively in \cref{tab:PalaciosNash,tab:OurPalaciosNash}, enabling us to conclude that payoffs, Nash probabilities, and empirical probabilities are all in agreement between \citeauthor{Palacios2003}'s results and our reproduction;
more quantitatively, the Jensen-Shannon divergence between \citeauthor{Palacios2003}'s results and ours is 0.84\% for the Nash distribution and 1.2\% for the empirical frequencies. 
We also notice that players' empirical action selection frequencies are quite close to the Nash-recommended frequencies, as measured by their Jensen-Shannon Divergence, and are actually playing an $\epsilon$-Nash equilibrium with a very low $\epsilon$ of $0.4 \%$.

\begin{table}[t]
    \centering
    \caption{Natural / Non-natural game restricted by footedness.}
    \label{tab:lcr_results}
    \begin{subtable}[b]{0.5\textwidth}
        \centering
        \caption{Left-footed players payoff table}
        \begin{tabular}{rRR}
            \toprule
            {} &       N-G & NN-G \EndTableHeader \\
            \midrule
            N-S  &  0.721 & 0.939 \\
            NN-S &  0.903 & 0.591 \\
            \bottomrule
        \end{tabular}
        \label{tab:lcr_left_footed}
    \end{subtable}%
    \begin{subtable}[b]{0.5\textwidth}
        \centering
        \caption{Right-footed players payoff table}
        \begin{tabular}{rRR}
            \toprule
            {} &       N-G & NN-G \EndTableHeader \\
            \midrule
            N-S  &  0.700 & 0.898 \\
            NN-S &  0.892 & 0.653 \\
            \bottomrule
        \end{tabular}
        \label{tab:lcr_right_footed}
    \end{subtable}%
\end{table}

\begin{table}[t]
    \centering
    \caption{Footedness equivalence p-value tables.}
    \begin{subtable}[b]{0.5\textwidth}
        \centering
        \caption{Natural / Non-natural game p-value table}
        \begin{tabular}{rMM}
        \toprule
        {} & N-G & NN-G \EndTableHeader \\
        \midrule
        N-S &  0.924566 &            0.170504 \\
        NN-S &  0.394900 &            0.407741 \\
        \bottomrule
        \end{tabular}
        \label{tab:nat_p_value}
    \end{subtable}%
    \begin{subtable}[b]{0.5\textwidth}
        \centering
        \caption{Left / Center / Right game p-value table}
        \begin{tabular}{rMMM}
            \toprule
            {} & R-G & C-G & L-G \EndTableHeader \\
            \midrule
            R-S &  0.000011 &  0.947369 &        6.931197e-01 \\
            C-S &  0.592054 &  0.868407 &        1.305657e-01 \\
            L-S &  0.017564 &  0.764020 &        7.791136e-07 \\
            \bottomrule
        \end{tabular}
        \label{tab:lcr_p_value}
    \end{subtable}%
    \par \bigskip
\end{table}

\begin{figure}[t]
    \centering
    \includegraphics[width=\textwidth]{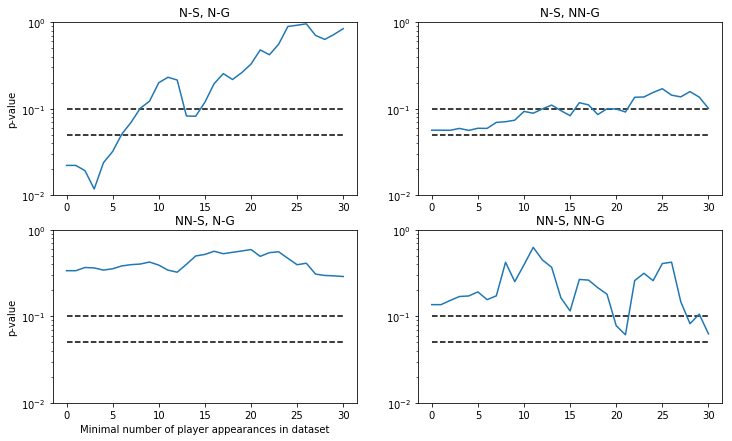}
    \caption{P-value table as a function of minimal experience.}
    \label{fig:p_value_by_exp}
\end{figure}

\begin{figure}[t]
    \centering
    \includegraphics[width=\textwidth]{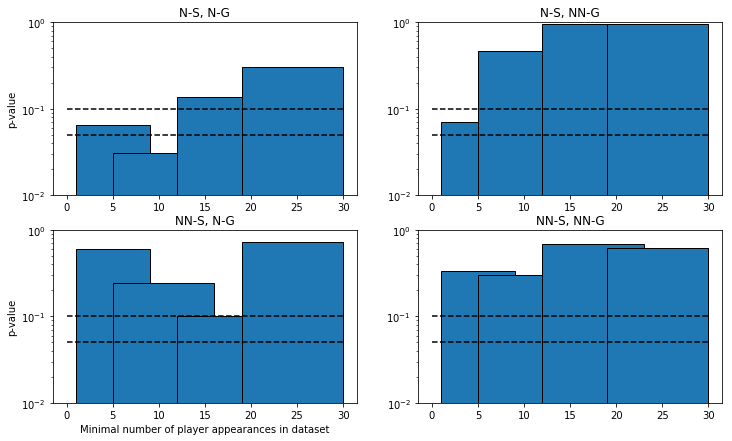}
    \caption{P-value table as a function of player-experience.}
    \label{fig:p_value_by_exp_bar}
\end{figure}

Having examined the similarity of payoff tables and distributions, we verify whether the Natural / Non-Natural game is statistically identical for left-footed and right-footed players (\cref{tab:lcr_results}), as assumed in \citet{Palacios2003}. 
To do so, we use a t-test to verify whether per-cell scoring rates are identical across footedness types.
The t-tests' p-values are reported in \cref{tab:nat_p_value}, and reveal that the games cannot be proven to be dissimilar across footedness and can, therefore, be assumed to be identical for left-footed and right-footed players. 
\cref{fig:p_value_by_exp} refines this result by representing the relationship between p-values of our t-test and minimal player appearance counts: when we modulate minimal appearance count of players in our test, the Natural Shot / Natural Goalkeeper cell goes from strongly dissimilar across footedness (low p-value) when including all players, to likely non-dissimilar (high p-value) when only including the players appearing the most in our dataset. 
This could be explained by low-appearance-counts-, which we take here as a proxy for low experience, kickers being less able to control their kicks, resulting in different control effectiveness for different footedness preferences, and in goalkeepers being less proficient in stopping shots going to their less frequently-kicked side (left) than to the other, a preference that we infer has been trained away in professional goalkeepers. 
To remove potential side-effects of merging data from low- and high-experience players together, \cref{fig:p_value_by_exp_bar} shows the relationship between p-values of our t-test and experience category where we allow for some overlap--between 1 and 7 shots, 5 and 12, etc.; the insight drawn from this figure is the same as that of \cref{fig:p_value_by_exp}, supporting the conclusion that experience removes the difference between left- and right-footed penalty kicks.

\begin{table}[t]
    \centering
    \caption{Left (L) - Center (C) - Right (R) tables for Shots (S) and Goalkeepers (G), with the three directions of kick/movement defined from the goalkeeper's perspective.}
    \begin{subtable}[b]{0.5\textwidth}
        \centering
    	\caption{Payoff table.}
    	\label{tab:lcr_table_noexp}
        \begin{tabular}{rRRR}
            \toprule
            {} &    R-G &    C-G & L-G \EndTableHeader \\
            \midrule
            R-S &  0.684 &  0.939 &            0.969 \\
            C-S &  0.964 &  0.160 &            0.953 \\
            L-S &  0.964 &  0.960 &            0.633 \\
            \bottomrule
        \end{tabular} 
    \end{subtable}%
    \par\bigskip
    \begin{subtable}[b]{1.0\textwidth}
        \centering
        \caption{Nash probabilities vs. Empirical frequencies corresponding to \subref{tab:lcr_table_noexp}.} \label{tab:lcr_table_noexp_nash}
        \begin{tabular}{rMMM|MMM}
            \toprule
            {} &       R-S &       C-S &       L-S &       R-G &       C-G & L-G \EndTableHeader \\
            \midrule
            Nash       &  0.478 &  0.116 &  0.406 &  0.441 &  0.178 &               0.381 \\
            Empirical  &  0.454 &  0.061 &  0.485 &  0.475 &  0.089 &               0.436 \\
            \bottomrule
        \end{tabular}
        \par
        \setlength{\fboxrule}{0pt}
        \fbox{Jensen–Shannon divergence: 0.75\%}
    \end{subtable}
\end{table}

We also analyzed the game defined by kicking to the left, center, or right, and confirmed \citeauthor{Palacios2003}'s intuition that it is fundamentally different across footedness preferences. 
Specifically, \cref{tab:lcr_table_noexp} synthesizes the empirical game corresponding to this new choice of actions, with aggregated scoring rates over both feet preferences. 
Note that in this case, left, center, and right are measured from the goalkeeper's perspective, such that the natural kick of a right-footed player would be considered a right kick. 
The  per-cell t-tests' p-values for this game are reported in \cref{tab:lcr_p_value}. Interestingly, the game is different when the goalkeeper jumps to the same side as the ball, but is otherwise mostly similar across footedness preference.
The empirical play frequencies for kickers, as reported in \cref{tab:lcr_table_noexp_nash}, are also further away from Nash frequencies than observed in the Natural / Non-Natural game (\cref{tab:OurPalaciosNash}), as can be seen from the Jensen-Shannon divergence between empirical frequencies and Nash (0.75\%, versus the the 0.087\% of the Natural / Non-Natural game) 
These insights indeed confirm the intuition that such a game is neither correct across footedness, nor the one the players follow.

Overall, these results provide insights into the impacts that the choice of actions have on conclusions drawn from empirical payoff tables. 
However, behavior and shooting styles also vary wildly per-player given footedness. 
If one is willing to consider several payoff tables (e.g., one per footedness), it seems natural to also take into account kickers' playing styles, as considered in the next section.

\subsection{Augmenting Game-theoretic Analysis of Penalty Kicks with Embeddings}

\begin{table}[t]
    \centering
    \caption{Cluster statistics.}
    \begin{tabular}{rrrrrr}
        \toprule
        {} & \# Players & \# Goals & \# Shots & Success rate (\%)   & Proportion of left-foot goals (\%) \\
        \midrule
        Cluster~1  &     197 &      144 &    167 & 86.2            &  10.4 \\
        Cluster~2  &     216 &       494 &   612 & 80.7            &  21.9\\
        Cluster~3  &     52 &         3 &     4 & 75.0             & 33.3\\
        Cluster~4  &     82  &       58 &    73 & 79.4             & 51.7 \\
        Cluster~5  &     87  &      44 &     60 & 73.3             & 34.1.0\\
        Cluster~6  &     1   &       0 &     0 & -               & 0.0 \\
        \midrule
        Total     &      635&      743 &    916  & 81.1            & 25.2 \\
        \bottomrule
    \end{tabular}
    \label{tab:cluster_stats}
\end{table}

\begin{table}[t]
    \centering
    \caption{Pair-wise comparison for the identified clusters. $<$ indicates that data was missing and minimum true p-value may be lower than the reported minimum p-value in the table. The symbol * indicates we cannot conclude whether clusters are different at the 10\% confidence level.}
    \begin{tabular}{p{5cm}rrrrrr}
        \toprule
        {}                                                                        & 1 vs. 2   & 1 vs. 4& 1 vs. 5   & 2 vs. 4   & 2 vs. 5    &  4 vs. 5 \\
        \midrule
        Min. cell $p$-value of t-test over table equality          & 4.49e-2  & $<$ 9.56e-2*  & $<$ 1.09e-1* & 4.49e-2  & 4.48e-2  & $<$ 3.39e-1*\\
        Jensen-Shannon divergence between Nash distr. (\%)                 & 0.03     & 0.57     & 0.09   & 0.35         & 0.02         & 0.21 \\
        Jensen-Shannon divergence between empirical distr. (\%)            & 0.06     & 0.01     & 0.06   & 0.08         & 0.24         & 0.04 \\
        Left footedness t-test $p$-value                                   & 3.43e-4  & 1.37e-7 & 3.18e-3 & 4.92e-5  & 1.07e-1 & 7.52e-2 \\
        \bottomrule
    \end{tabular}
    \label{tab:cluster_stat_test}
\end{table}

\begin{table}[t]
    \centering
    \caption{p-values for t-test that empirical action distributions are equal among different clusters. Minimum p-value (across kicker and goalkeeper roles) is indicated in bold for each row.}
    \begin{tabular}{rrr}
    	\toprule
        Kicker clusters compared &  Kicker p-value  &  Goalkeeper p-value  \\
        \midrule
    	1 vs. 2  &  0.52  &  \textbf{0.05}\\
    	1 vs. 4  &  \textbf{0.85}  &  0.95\\
    	1 vs. 5  &  0.42  &  \textbf{0.27}\\
    	2 vs. 4  &  0.52  &  \textbf{0.14}\\
    	2 vs. 5  &  0.51  &  \textbf{0.16}\\
    	4 vs. 5  &  0.4  &  \textbf{0.26}\\
    	\bottomrule
    \end{tabular}
    \label{tab:empirical_p_value_table}
\end{table}

\begin{figure}[t]
    \centering
    \begin{subfigure}[b]{0.48\textwidth}
        \centering
    	\includegraphics[width=\textwidth]{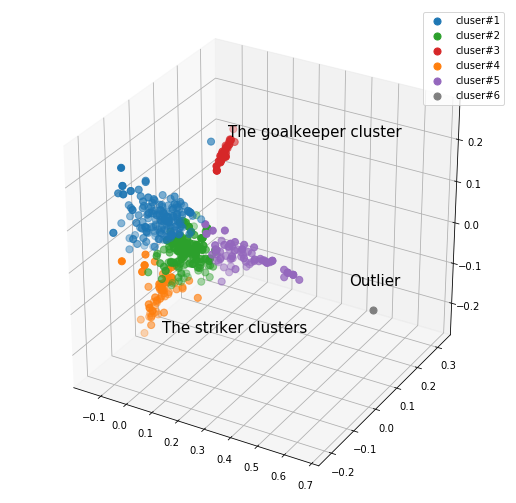}
    	\caption{}
        \label{fig:striker_goalie_clusters}
    \end{subfigure}%
    \hfill
    \begin{subfigure}[b]{0.48\textwidth}
        \centering
        \includegraphics[width=\textwidth]{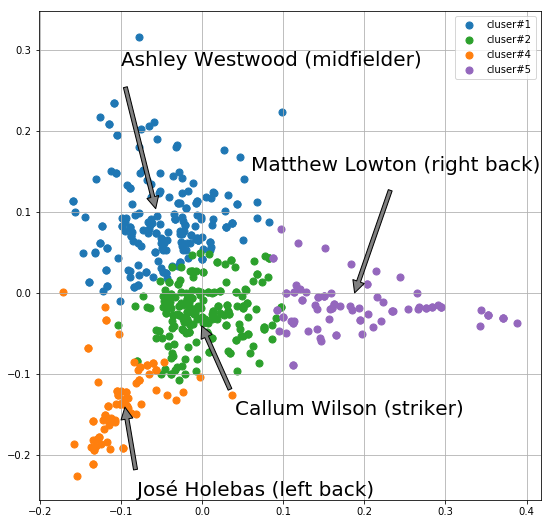}
    	\caption{}
        \label{fig:striker_clusters}
    \end{subfigure}\\
    \caption{Visualization of the identified player clusters. \subref{fig:striker_goalie_clusters} visualizes the goalkeeper cluster, the kicker clusters and an outlier automatically detected through K-means clustering. To show the separation of the kicker clusters clearly, we visualize them in \subref{fig:striker_clusters} after removing the goalkeeper and outlier clusters, and we also label each cluster with a Premier League player in it.
    }
\end{figure}

\begin{table}[t]
    \centering
    \caption{Nash probabilities and empirical frequencies tables for Shot (S) and Goalkeepers (G) with Natural (N) and Non-Natural (NN) actions. Note that Cluster 3 is omitted due to it consisting of very few shots (taken by goalkeepers).}
    \label{tab:egta_pv_nash_emp}
    \begin{subtable}[b]{0.5\textwidth}
        \centering
    	\caption{All players. 916 total shots.}
            \begin{tabular}{rMM|MM}
            \toprule
            {} &     NN-S &    N-S & NN-G & N-G \EndTableHeader \\
            \midrule
            Nash       &  0.391 &  0.609 &  0.406 &                0.594 \\
            Empirical  &  0.503 &  0.497 &  0.413 &                0.587 \\
            \bottomrule
            \end{tabular}
            \setlength{\fboxrule}{0pt}
            \fbox{$\epsilon$-Nash equilibrium: $\epsilon=2.71\%$}
        \label{tab:egta_pv_all}
        \\
    \end{subtable}%
    \begin{subtable}[b]{0.5\textwidth}
        \centering
        \caption{Kickers in Cluster~1. 167 total shots.}
            \begin{tabular}{rMM|MM}
            \toprule
            {} &     NN-S &    N-S & NN-G & N-G \EndTableHeader \\
            \midrule
            Nash &  0.423  &  0.577  &  0.379  &  0.621 \\
            Empirical &  0.485 & 0.515 & 0.371 & 0.629  \\
            \bottomrule
            \end{tabular}
            \setlength{\fboxrule}{0pt}
            \fbox{$\epsilon$-Nash equilibrium: $\epsilon=0.08\%$}
        \label{tab:egta_pv_c1}
        \\
    \end{subtable}\\
    \par\bigskip
    \begin{subtable}[b]{0.5\textwidth}
        \centering
        \caption{Kickers in Cluster~2. 612 total shots.}
        \begin{tabular}{rMM|MM}
            \toprule
            {} &     NN-S &    N-S & NN-G & N-G \EndTableHeader \\
            \midrule
            Nash &  0.401  &  0.599  &  0.430  &  0.570 \\
            Empirical &  0.520 & 0.480 & 0.418 & 0.582  \\
            \bottomrule
        \end{tabular}
        \setlength{\fboxrule}{0pt}
        \fbox{$\epsilon$-Nash equilibrium: $\epsilon=2.89\%$}
        \label{tab:egta_pv_c2}
    \end{subtable}%
    \begin{subtable}[b]{0.5\textwidth}
        \centering
        \caption{Kickers in Cluster~4. 73 total shots.}
            \begin{tabular}{rMM|MM}
            \toprule
            {} &     NN-S &    N-S & NN-G & N-G \EndTableHeader \\
            \midrule
            Nash &  0.320  &  0.680  &  0.375  &  0.625 \\
            Empirical &  0.479 & 0.521 & 0.438 & 0.562  \\
            \bottomrule
            \end{tabular}
            \setlength{\fboxrule}{0pt}
            \fbox{$\epsilon$-Nash equilibrium: $\epsilon=5.17\%$}
        \label{tab:egta_pv_c4}
    \end{subtable}\\
    \par\bigskip
    \begin{subtable}[b]{0.5\textwidth}
        \centering
        \caption{Kickers in Cluster~5. 60 total shots.}
            \begin{tabular}{rMM|MM}
            \toprule
            {} &     NN-S &    N-S & NN-G & N-G \EndTableHeader \\
            \midrule
            Nash &  0.383  &  0.617  &  0.317  &  0.683 \\
            Empirical &  0.450 & 0.550 & 0.400 & 0.600  \\
            \bottomrule
            \end{tabular}
            \setlength{\fboxrule}{0pt}
            \fbox{$\epsilon$-Nash equilibrium: $\epsilon=4.86\%$}
        \label{tab:egta_pv_c5}
    \end{subtable}
\end{table}

\begin{figure}[t]
    \centering
    \includegraphics[width=\textwidth]{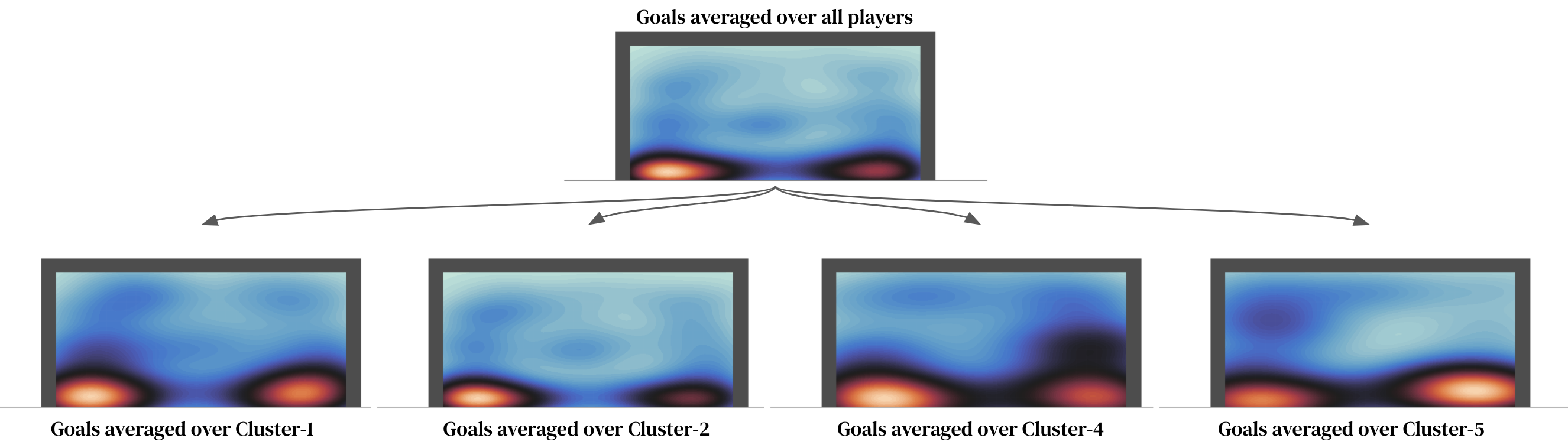}
    \caption{Heatmaps of goals by all kickers and kickers in individual clusters with respect to empirical probabilities. We exclude the goalkeeper cluster (Cluster~3) and the outlier cluster (Cluster~6) because of insufficient samples.
    }
    \label{fig:egta_pv_cluster_heatmap}
\end{figure}

While the previous section undertook a descriptive view of the penalty kick scenario (i.e., providing a high-level understanding of kicker and goalkeeper play probabilities), here we investigate whether we can find the best strategy for a player given the knowledge of the kicker's play style. 
In game-theoretic terms, we conduct a prescriptive analysis of penalty kicks to enable informed decision-making for players and coaching staff in specific penalty kick situations.
Ideally, one would iterate the earlier empirical payoff analysis for every possible combination of goalkeeper and kicker in a given league, thus enabling decision-making at the most granular level; 
however, the inherent sparsity of penalty kick data makes such an approach infeasible.
Instead, we introduce a meaningful compromise here by combining statistical learning with game theory (i.e., \frontierOneGTSL), first quantifying individual playing styles, then using clustering techniques to aggregate players (i.e., both strikers and goalkeepers) based on said styles, and finally synthesizing empirical games for each identified cluster.
We focus our analysis on penalties including all players who participated in Premier League matches from 2016 to 2019.

On a technical level, our approach consists of the three following steps.
First, we characterize the playing style of a player in a manner that can be interpreted both by human experts and machine learning systems.
In particular, we use Player Vectors~\citep{DecroosD19} to summarize the playing styles of kickers using an 18-dimensional real-valued vector. 
These Player Vectors are extracted from historical playing trajectories in real matches, with technical details provided in \cref{sec:player_vectors}.
Each dimension of the Player Vector corresponds to individual on-pitch player behaviors (e.g., styles of passes, take-ons, shots, etc.), and the value of each dimension is standardized and quantifies the weight of that particular action style for the considered player. We also filter experienced players with at least 50 appearances in the Premier League matches from 2016 to 2019.
In total, we obtain 635 such vectors for the individual players in our dataset.
Second, we cluster players in accordance to their Player Vectors, using K-means with the number of clusters chosen as the value causing the most significant drop in inertia (a standard heuristic). 
This process yields 6 clusters in total, with statistics summarized in \cref{tab:cluster_stats}.
In particular, K-means clustering detects an outlier cluster with only one player (Cluster~6), and we also observe that there are very few shot samples in Cluster~3, as it consists of a cluster of goalkeepers (an interesting artifact illustrating the ability of Player Vectors and K-means clustering to discern player roles).
Given the few samples associated with these two clusters, we henceforth exclude them from the game-theoretic analysis. 
We observe that cluster pairs (1, 2), (1, 4), (2, 4), and (2, 5) are significantly different, with the minimum cell-wise p-values for these cluster pairs smaller than 0.10 in~\cref{tab:cluster_stat_test}. 
We therefore focus our game-theoretic analysis on these cluster pairs.
Moreover, we also qualitatively illustrate differences between the clusters in \cref{fig:striker_goalie_clusters,fig:striker_clusters}, which visualize the results of reducing the Player Vectors dimensionality from 18 to, respectively, 3 and 2 via Principal Component Analysis.
Here, we observe that the goalkeeper cluster is well-separated from the kicker clusters in \cref{fig:striker_goalie_clusters}, and in order to better visualize the kicker clusters, we project \cref{fig:striker_goalie_clusters} onto its x and y axis after removing the goalkeeper and outlier clusters in \cref{fig:striker_clusters}.
We also identify therein the most representative kicker per-cluster (i.e., the player whose feature vector is closest to the mean of the corresponding cluster)

Finally, we conduct the aforementioned game-theoretic analysis for each cluster.
In our earlier \cref{tab:cluster_stats}, we observe that the kickers in some clusters have different success rates in penalty kicks.
Moreover, a closer behavioral analysis yields deeper insights.
We first examine the Nash strategies played by each cluster, and then visualize the actual play behavior with respect to empirical probabilities in \cref{fig:egta_pv_cluster_heatmap}.
\Cref{tab:egta_pv_all} summarizes the overall Nash distributions for all players considered, with \cref{tab:egta_pv_c1,tab:egta_pv_c2,tab:egta_pv_c4,tab:egta_pv_c5} showing cluster-specific distributions.
These tables illustrate that the kickers have the same empirical behavior, an assertion statistically confirmed in \cref{tab:empirical_p_value_table}; yet their Nash-derived recommendations are different: although kickers in all clusters are recommended by the Nash to shoot more to their natural sides than to their non-natural sides, the recommended strategy for kickers in Cluster~1 is actually quite balanced between natural and non-natural shots.
This greater imbalance is shown by comparing Jensen-Shannon divergence. As we see in \cref{tab:cluster_stat_test}
, the Jensen-Shannon divergence of the Nash probabilities between Cluster~1 and 4 (0.57\%) is 6-7 times greater than that between Cluster~1 and 5 (0.09\%) and 19 times greater than that between Cluster~1 and 2 (0.03\%).
We also notice that the clusters' players are all playing epsilon Nash equilibra with relatively low epsilon (\cref{tab:egta_pv_nash_emp}).
In other words, although their empirical strategies seem to deviate from corresponding Nash strategies action-wise,
the expected payoffs of these two strategies are close, and they could still stand to gain in "stability" by switching to corresponding Nash strategy.
Nevertheless, most of these Nash recommendations come from very low-sample empirical payoff tables, which entails potentially inaccurate Nash distributions. 
We nevertheless note that this low-data regime is induced by the restriction of our analysis to players having played in matches of Premier League only from 2016 to 2019.
Obtaining Player Vector data for all players in our dataset would allow us to study cluster behavior with greater statistical precision. 
Nevertheless, the current study leaves no statistical doubt regarding the pertinence of clustering payoff tables using player embeddings--specifically Player Vectors.

Qualitatively, in addition to analyzing the strategies with respect to Nash probabilities, the patterns of positions of the ball of successful goals also vary from clusters to clusters, as visualized in \cref{fig:egta_pv_cluster_heatmap}. For instance, kickers in Cluster~2 tend to score mostly to the bottom left corner of the goalmouth, while the scoring positions in other clusters are more balanced, though these could also be partly due to lower sample sizes for some clusters.

\begin{figure}[t]
    \centering
    \begin{subfigure}[b]{0.48\textwidth}
        \centering
    	\includegraphics[width=\textwidth]{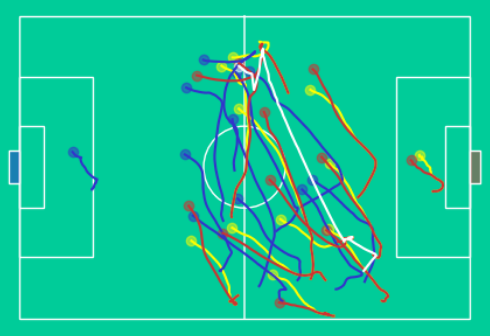}
    	\caption{}
        \label{fig:ghosting_before}
    \end{subfigure}%
    \hfill
    \begin{subfigure}[b]{0.48\textwidth}
        \centering
        \includegraphics[width=\textwidth]{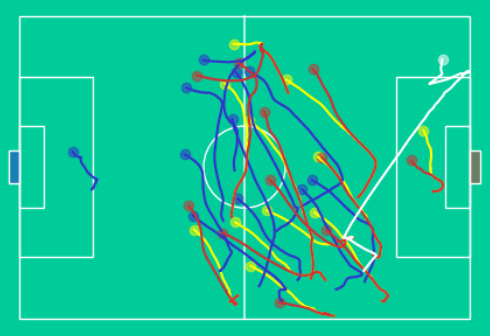}
    	\caption{}
        \label{fig:ghosting_after}
    \end{subfigure}\\
    \begin{subfigure}[b]{\textwidth}
        \centering
        \begin{tabular}{cccc}
         \tikzcircle[black, fill=white]{4.0pt} Ball (truth) &  \tikzcircle[white, fill=blue]{4.0pt} Attackers (truth) &  \tikzcircle[white, fill=red]{4.0pt} Defenders (truth) &  \tikzcircle[white, fill=yellow]{4.0pt} Defenders (predicted)
        \end{tabular}
    \end{subfigure}
    \caption{Predictive modeling using football tracking data. \subref{fig:ghosting_before} visualizes predictions under the original data. Here, ground truth information for all players and the ball is provided to the underlying predictive model, with defender positions truncated and predicted by the model after a cut-off time (as indicated by the yellow traces). \subref{fig:ghosting_after} illustrates the same scenario, after counterfactual perturbation of the ground truth ball direction to ascertain the predicted reaction of the defending goalkeeper (far right).
    }
    \label{fig:ghosting}
\end{figure}

\subsection{Generative Trajectory Prediction Models for Counterfactual Analysis}\label{sec:predictive_models_counter}

Ghosting refers to the prescription of the trajectories the players in a sports team should have executed, in contrast to what they actually did~\citep{lowe_lights_2013}.
Solution of this and the broader problem class of generative trajectory prediction implies benefits spanning from recommendation of trajectories or setups for constrained set pieces, then to short-term plays involving a subset of players, and eventually to long-term strategies/plays for the entire team. 
Team-level predictions would also strongly benefit from game-theoretic and multi-agent considerations, and is perceived to play a key role in an established AVAC system. 
We here present an illustrative example to ground the earlier discussion regarding the potential impacts of using learned predictive models to conduct counterfactual analysis of football matches.

For example, one might train a trajectory prediction model on league data (e.g., as done in \citet{LeY0L17}), provide an input context to such a model (e.g., consisting of the true state of the ball, defenders, and attackers up to some point in time), and subsequently predict future trajectories of players.
\Cref{fig:ghosting_before} visualizes league-average predicted behaviors conditioned on such an input context.
This illustrative example was trained using a baseline predictive model, similar to that of \citet{Le17}.
Here we trained a centralized long-short term memory model (of 2 layers, each with 256 units), taking as input the raw trajectories of players and the ball, and predicting as output the step-wise change in trajectory of the defensive players.
The model was trained on 240 frames of 25~fps tracking data, downsampled to 12.5~fps, with half the frames in each play used for providing a prediction context, and the other half occurring at the prediction cut-off. 
We used the $l_2$-loss on the tracking data for training, and randomized the order of attacking and defending players to avoid the role-assignment problem mentioned in \citet{Le17} (similar to one of the baseline approaches of \citet{yeh2019diverse}). 

As pointed out in the literature~\citep{Le17,LeY0L17,yeh2019diverse,li2020generative}, a key advantage of generative predictive models is that they can be used for counterfactual analysis of play outcomes.
We illustrate such an example in \cref{fig:ghosting_after}, where we perturb the trajectory of the ball, inferring the subsequent behaviors of defenders in reaction (noting, e.g., the tendency of the goalkeeper to chase the ball in reaction to it entering the penalty area).
While simple, case-by-case counterfactual case studies such as the above have been conducted to some extent in the literature, consideration of responses to more complex perturbations (e.g., changes of one team's tactics or meta-strategy as a whole, changes in player behavior due to injuries, or changes due to substitutions of individual players) bear potential for significantly more in-depth analysis. 

\section{Discussion}

Football analytics poses a key opportunity for AI research that impacts the real world.
The balance of its reasonably well-controlled nature (versus other physical domains beyond sports, e.g., search-and-rescue), considerations associated with human factors (e.g., heterogeneous skill sets, physiological characteristics such as injury risks for players, etc.), and the long-term cause-and-effect feedback loop due to the relative infrequency of scoring even in professional play make it a uniquely challenging domain.
Nonetheless, the rapidly-emerging availability of multi-modal sensory data make it an ideal platform for development and evaluation of key AI algorithms, particularly at the intersection of the aforementioned fields of statistical learning, computer vision, and game theory. 

In this paper, we highlighted three frontiers at the intersection of the above fields, targeting the simultaneous advancement of AI and football analytics. 
We highlighted the overlying goal of developing an Automated Video Assistant Coach (AVAC), a system capable of processing raw broadcast video footage and accordingly advising coaching staff in pre-, in-, and post-match scenarios.
We subsequently illustrated how the combination of game theory and statistical learning could be used to advance classical results in football analytics, with an in-depth case study using a dataset comprised of over 15000 penalty kicks, and subsequently combined with the Player Vectors analysis of~\citet{DecroosD19} to discern kicking styles.

\begin{figure}[t!]
    \centering
    \includegraphics[width=\textwidth]{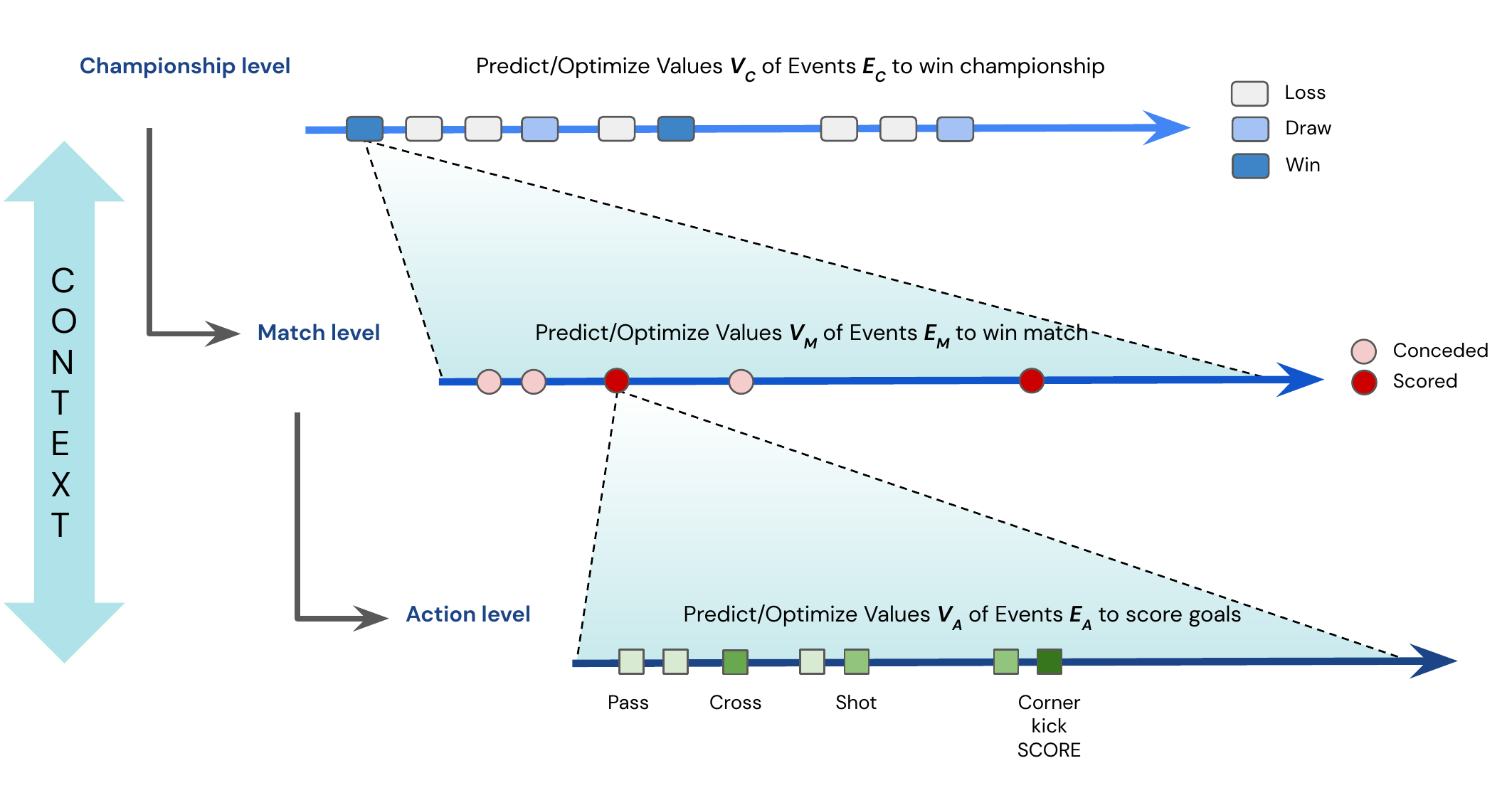}
    \caption{A multi-level view of football analytics cast as a reinforcement learning problem. We discern three levels: the top level aims to learn how to win championships by winning matches; the middle level optimizes for winning a match; finally, the bottom level seeks to optimize goal-scoring. The context between these various level is shared in both a top-down and bottom-up fashion. }
    \label{fig:RLview}
\end{figure}

A notable observation for future work focusing on prescriptive football analytics is that the domain and some of the state-of-the-art research bear key similarities to RL. 
At a high level, the process of winning football championships can be cast as a sequential decision-making problem, with a concrete reward structure centered on three timescales of increasing abstraction:
scoring goals, winning matches, and subsequently winning championships. We illustrate this view in \cref{fig:RLview}.
Under this hierarchical view of football, each layer can be considered an RL problem at the designated level of abstraction.
For example, at the lowest level, the sequential decisions made by teammates that lead to a goal can be considered a policy mapping states to actions, using the lexicon of RL.
Likewise, estimates of the value of player actions based on the outcomes associated with actions taken in real games (as in VAEP~\citep{decroos2019actions}) can be considered analogous to those that learn action-values associated with RL policies.
Further expanding this analogy, learning to quantify the contribution of individual players to a team's estimated goal-scoring value can be cast as a so-called credit assignment problem, a key area of research in RL.
Finally, given the presence of multiple on-pitch players with both cooperative and competitive incentives, the value function learning problem situates itself in the area of multi-agent RL. 
Multi-agent RL, critically, seeks to understand and learn optimal policies for agents in such interactive environments, linking also to game theory in providing the appropriate mathematical foundations to model this strategic process.
As such, the multi-agent RL approach fits well under \frontierOneGTSL, which considers the game-theoretic interactions of strategic players given specified payoffs, and use of learning techniques for identifying optimal policies. 
Moreover, this connection also highlights a potential overlap of interest between real-world football and RoboCup, in that the RL paradigm can be used to optimize player and robot policies alike, despite the widely-different player embodiments considered in each of these two fields.
Overall, such parallels can be drawn at all levels of abstraction highlighted in the aforementioned hierarchical process modeling football championships, implying the foreseeable importance of the RL paradigm as football analytics shifts from understanding the game to subsequently optimizing player and team decisions at increasingly broader levels. 

Moreover, the toolkits developed within the context of football analytics are also likely to have direct benefits for closely-related fields, and could be foreseeably adapted to many other sports.
One interesting extension concerns the application of football analytics techniques to the emerging field of eSports, wherein there is a large amount of data collected (in both raw video form, and structured data formats), e.g., such data streams are available for games such as Dota~2 or StarCraft.
In Dota~2, for example, a coaching functionality analogous to that in football is available, wherein an experienced player is connected to the game and advises other players on various strategic tactics.
Moreover, several of the most popular eSports games are inherently multi-player, in the sense that their outcomes are not determined by only an individual's skill, but a team's skill, mixing cooperative and competitive behaviors (as in football).
Automatic analysis of games could provide insights into weak and strong points of teams, tactics used,  and directions for improvement.
These related domains could, therefore, provide a low-hanging fruit for football analytics techniques to generalize, in a seamless manner, beyond football.

Overall, the combination of data sources, downstream benefits on related domains, and potentials for impact that AI could have on the football domain are quite evident.
Perhaps more importantly, the promising commensurate impacts of football analytics on AI research (through the feedback loop established between the football microcosm to the three foundational fields highlighted in \cref{fig:overview}) are foreseen to make football a highly appealing domain for AI research in coming years.   

\section*{Acknowledgments}
The authors gratefully thank Thomas Anthony and Murray Shanahan for their helpful feedback during the paper writing process. 

\begin{appendices}
\crefalias{section}{appendix}

\section{Additional Works Related to Statistical Learning in Football}\label{sec:additional_sl_works}

Evaluating the effect of individual actions throughout the game is challenging as they naturally depend on the circumstances in which they were performed and have long-term consequences that depend on how the sequence plays out. Most works have focused on measuring the quality of specific action types in distinct concrete game situations~\citep{barr2008evaluating,spearman2018beyond,bransen2018measuring}.
More recent work has focused on a unifying view in which actions are valued according to how they increase or decrease the likelihood of the play leading to a goal~\citep{decroos2019actions,Fernandez2019}.
The main idea is to estimate the value of a given `state' of the game. Intuitively, the state of a particular game includes everything that happened in the match until this point, including the score, identities of players and associated traits, time left on the clock, all prior actions, position of the players and the ball, etc.; moreover, one may wish to also consider the state of a tournament as a whole (e.g., previous and upcoming matches, the number of yellow cards accrued by players, etc.). 
A recent method used for assigning values to on-ball actions is known as \textit{Valuing Actions by Estimating
Probabilities} (VAEP) \citep{decroos2019actions}. 
Actions are valued by measuring their effect on the game state and in turn the probabilities that a team will score. These scores can then be used to assess contribution of players to a team or measuring the mutual chemistry for a pair of players \citep{bransen2020chemistry}.

Finally, a promising application of statistical learning is the development of models that can carry out temporal predictions. 
This area  is closely related to trajectory prediction \cite{wang2007gaussian,gupta2018social,fernando2018gd,deo2018convolutional,alahi2016social}.
In the context of sports analytics, such trajectory prediction models can be useful for conducting the form of analysis known as \emph{ghosting}, which, given a particular play, predicts the actions that a \emph{different} team or player would have executed. 
Beyond just capturing game dynamics, models that can accurately carry out predictions could constitute valuable tools for counterfactual reasoning, which allows us to consider the outcomes of alternative scenarios that never actually took place.
So far, such predictive models have been primarily used for predicting the trajectory of the ball~\citep{maksai2016players} and of players themselves~\citep{Le17, LeY0L17,su2019graph,li2020generative,yeh2019diverse}. 
Also of importance are models which identify player roles from predicted trajectories~\citep{FelsenLG18}.

\section{Pose Estimation}\label{sec:pose_estimation}
As previously illustrated, multi-person human pose estimation \citep{pavlakos2017coarse,pavlakos2019expressive,he2020epipolar,pavllo20193d,lassner2017unite,Cheng_2019_ICCV,iskakov2019learnable} is a central part of vision-based analysis of football video.
Methods for this task can be grouped into two types: one the one hand, bottom-up approaches first detect human joints, and group them into pose instances~\citep{iqbal2016multi,fang2017rmpe,papandreou2017towards,huang2017coarse,he2017mask,sun2019deep}; 
on the other, top-down approaches first detect body instances and run single-person pose estimation models on each instance~\citep{pishchulin2016deepcut,insafutdinov2016deepercut,cao2017realtime,newell2017associative,papandreou2018personlab,kocabas2018multiposenet}.
The computation cost of top-down methods increases linearly with the number of people in an image, while that of bottom-up methods stays constant.
However, in cases where there is significant overlap between instances, top-down approaches are often more accurate~\citep{chen2020monocular}.

We experimented with G-RMI~\citep{papandreou2017towards}, a well-established top-down approach, and give examples of predictions in \cref{fig:poses_penalty_kick}.
In the first stage, Faster-RNN~\citep{ren2015faster} is used to detect person instances.
Inspired by detection methods, the second stage combines classification and regression to process each resulting crop:
a fully convolutional network first densely classifies whether each spatial position is in the vicinity of a given keypoint class, and then refines each prediction by predicting an offset.
A specialized form of Hough voting (see \citep{duda1972use} for background) is introduced to aggregate these predictions and form highly localized activation maps.
A key-point based confidence score and non-maximum suppression procedure further improve results.
We plan to build on this approach to develop methods for the previously mentioned challenges.

\section{Player Vectors}\label{sec:player_vectors}
In particular, we follow definition of \textit{playing style} in~\citet{DecroosD19}, which is defined as a player's preferred area(s) on the field to occupy and which actions they tend to perform in each of these locations, and generate our player vectors with the method proposed in~\citet{DecroosD19}. 
The procedure of generating player vectors unfolds into four steps.
First, we collect the event stream data of all Premier League matches that Liverpool Football Club participated in from 2017 to 2019, and filter the actions of types passes, dribbles, shots and crosses. 
Secondly, for each pair of player $p$, who is observed in the event stream dataset, and relevant action type $t$, we overlay a grid of size $60 \times 40$ on the football pitch and count how many times player $p$ performed action $t$ in each grid cell. This procedure yields a matrix which summarizes spatial preference of player $p$ performing action type $t$.
Thirdly, we compress that matrix into a small vector. To do this, we reshape each matrix into a vector and group it together with all other vectors of the same action type, and we then perform non-negative matrix (NMF) factorization to reduce the dimensionality of these matrices. This procedure yields a smaller vector, and the value of each dimension quantifies the preference of player $p$ performing the action type $t$ in the area $a$.
Finally, for each player, we obtain 4 vectors corresponding to the 4 action types, and we generate one final vector of 18 dimensions by concatenating his compressed vectors for relevant action types.

\end{appendices}

\vskip 0.2in
\bibliography{References}
\bibliographystyle{apacite}

\end{document}


\maketitle

\paragraph{Additional works related to statistical learning in football}

Evaluating the effect of individual actions throughout the game is challenging as they naturally depend on the circumstances in which they were performed and have long-term consequences that depend on how the sequence plays out. Most works have focused on measuring the quality of specific action types in distinct concrete game situations.\citep{barr2008evaluating,spearman2018beyond,bransen2018measuring} More recent work has focused on a unifying view in which actions are valued according to how they increase or decrease the likelihood of the play leading to a goal.\citep{decroos2019actions,Fernandez2019}
The main idea is to estimate the value of a given `state' of the game. Intuitively, the state of a particular game includes everything that happened in the match until this point, including the score, identities of players and associated traits, time left on the clock, all prior actions, position of the players and the ball, etc.; moreover, one may wish to also consider the state of a tournament as a whole (e.g., previous and upcoming matches, the number of yellow cards accrued by players, etc.). 
A recent method used for assigning values to on-ball actions is known as \textit{Valuing Actions by Estimating
Probabilities} (VAEP) \citep{decroos2019actions}. 
Actions are valued by measuring their effect on the game state and in turn the probabilities that a team will score. These scores can then be used to assess contribution of players to a team or measuring the mutual chemistry for a pair of players. \citep{bransen2020chemistry}

Finally, a promising application of statistical learning is the development of models that can carry out temporal predictions. This area  is closely related to rajectory prediction.\cite{wang2007gaussian,gupta2018social,fernando2018gd,deo2018convolutional,alahi2016social} 
In the context of sports analytics, such trajectory prediction models can be useful for conducting the form of analysis known as \emph{ghosting}, which, given a particular play, predicts the actions that a \emph{different} team or player would have executed. 
Beyond just capturing game dynamics, models that can accurately carry out predictions could constitute valuable tools for counterfactual reasoning, which allows us to consider the outcomes of alternative scenarios that never actually took place.
So far, such predictive models have been primarily used for predicting the trajectory of the ball~\citep{maksai2016players} and of players themselves~\citep{Le17, LeY0L17,su2019graph,li2020generative,yeh2019diverse}. 
Also of importance are models which identify player roles from predicted trajectories~\citep{FelsenLG18}. 

\section{Dataset}

\so{Potentially move to supp info?}\kt{yes agreed}

%% file: imports.tex
\usepackage{xspace}
\usepackage[utf8]{inputenc}
\usepackage[hidelinks]{hyperref}
\usepackage{subcaption}
\usepackage{jair}
\usepackage[natbibapa]{apacite} 
\usepackage[title,toc,titletoc,page]{appendix}

\usepackage{tabularx}
\usepackage{graphicx}
\usepackage{natbib}
\usepackage{multirow}
\usepackage{nicefrac}
\usepackage[margin=1in]{geometry}
\usepackage{xspace}
\usepackage{nameref}
\usepackage{lineno}
\usepackage{amsmath}
\usepackage{amssymb}
\usepackage{comment}
\usepackage{nameref}
\makeatletter
\AtBeginDocument{%
  \@ifdefinable{\myorg@nameref}{%
    \LetLtxMacro\myorg@nameref\nameref
    \DeclareRobustCommand*{\nameref}[1]{%
      \emph{\myorg@nameref{#1}}%
    }%
  }%
}
\makeatother

\usepackage[table]{xcolor}
\usepackage{pgf}
\usepackage{collcell}
\usepackage{booktabs}

\usepackage{etoolbox}

\newtoggle{inTableHeader}
\toggletrue{inTableHeader}
\newcommand*{\StartTableHeader}{\global\toggletrue{inTableHeader}}%
\newcommand*{\EndTableHeader}{\global\togglefalse{inTableHeader}}%

\newtheorem{theorem}{Theorem}
\newtheorem{definition}[theorem]{Definition}
\let\OldTabular\tabular%
\let\OldEndTabular\endtabular%
\renewenvironment{tabular}{\StartTableHeader\OldTabular}{\OldEndTabular\StartTableHeader}%

\newcommand*{\MinNumber}{0.0}%
\newcommand*{\MidNumber}{0.5} %
\newcommand*{\MaxNumber}{1.0}%
\definecolor{egtaRed}{HTML}{d61f0b}
\definecolor{egtaYellow}{HTML}{fcf560}
\definecolor{egtaGreen}{HTML}{0ec74f}
\definecolor{egtaBlue}{HTML}{34b1eb}

\newcommand{\ApplyGradient}[1]{%
  \iftoggle{inTableHeader}{#1}{
    \ifdim #1 pt > \MidNumber pt
        \pgfmathsetmacro{\PercentColor}{max(min(100.0*(#1 - \MidNumber)/(\MaxNumber-\MidNumber),100.0),0.00)} %
          \edef\x{\noexpand\cellcolor{egtaGreen!\PercentColor!white}}\x\textcolor{black}{#1}%
    \else
        \pgfmathsetmacro{\PercentColor}{max(min(100.0*(\MidNumber - #1)/(\MidNumber-\MinNumber),100.0),0.00)} %
      \edef\x{\noexpand\cellcolor{egtaRed!\PercentColor!white}}\x\textcolor{black}{#1}%
    \fi
  }}
\newcolumntype{R}{>{\collectcell\ApplyGradient}r<{\endcollectcell}}

\newcommand{\ApplyGradientMono}[1]{%
  \iftoggle{inTableHeader}{#1}{
    \pgfmathsetmacro{\PercentColor}{max(min(100.0*(#1 - \MinNumber)/(\MaxNumber-\MinNumber),100.0),0.00)} %
      \edef\x{\noexpand\cellcolor{egtaBlue!\PercentColor!white}}\x\textcolor{black}{#1}%
  }}
  
\newcommand{\gamematrix}[6]{
  \begin{tabular}{c} 
    $\begin{array}{c} \\ #1 \\ #2 \end{array} \hspace{0mm}
    \begin{array}{r@{}c@{}c@{}l}
      & #1 & #2 & \\
      \bigg( & \begin{array}{c} #3 \\ #5 \end{array} & \begin{array}{c} #4 \\ #6 \end{array} & \bigg)
    \end{array}$
  \end{tabular}}
  
\newcolumntype{M}{>{\collectcell\ApplyGradientMono}r<{\endcollectcell}}

\usepackage{tikz}
\newcommand{\tikzcircle}[2][red,fill=red]{\tikz[baseline=-0.8ex]\draw[#1,radius=#2] (0,0) circle ;}%

\input{math_commands.tex}

\usepackage[capitalise,noabbrev]{cleveref} 
\crefname{equation}{}{} 

\newcommand{\frontierOneGTSL}{Frontier~1~(GT\&SL)\xspace}
\newcommand{\frontierTwoSLV}{Frontier~2~(SL\&CV)\xspace}
\newcommand{\frontierThreeGTV}{Frontier~3~(GT\&CV)\xspace}

%% file: math_commands.tex
\usepackage{amsmath,amsfonts,bm}

\def\eqref#1{equation~\ref{#1}}

\def\1{\bm{1}}

\DeclareMathAlphabet{\mathsfit}{\encodingdefault}{\sfdefault}{m}{sl}
\SetMathAlphabet{\mathsfit}{bold}{\encodingdefault}{\sfdefault}{bx}{n}

\newcommand{\E}{\mathbb{E}}

